\pdfoutput=1
\documentclass[11pt]{article}
\usepackage{hyperref}
\usepackage{url}
\usepackage{xurl}
\usepackage{graphicx}
\usepackage{booktabs}
\usepackage{amssymb}
\usepackage{amsmath}
\usepackage{subcaption}
\usepackage{xcolor}
\usepackage{adjustbox} 
\usepackage{multirow}
\usepackage{wrapfig}
\usepackage{tikz}
\usepackage[inline]{enumitem}
\usepackage{xspace}
\usepackage{soul}
\newcommand{\etal}{\textit{et al}.}

\newcommand{\eg}{\textit{e}.\textit{g}., }
\usepackage{titletoc}

\definecolor{bg_blue}{RGB}{213,227,251}

\definecolor{bg_yellow}{RGB}{250,243,187}
\definecolor{bg_purple}{RGB}{177,167,207}
\definecolor{bg_red}{RGB}{200,169,188}
\definecolor{bg_green}{RGB}{192,213,175}
\definecolor{bg_skin}{RGB}{245,232,210}

\definecolor{red_color}{RGB}{255,0,0}

\definecolor{yellow_color}{RGB}{255,202,47}

\definecolor{purple_color}{RGB}{64,103,139}

\definecolor{dark_red}{RGB}{153, 31, 41}

\definecolor{green_color}{RGB}{130,139,78}

\definecolor{brown_color}{RGB}{205,90,161}

\definecolor{lg_color}{RGB}{63,147,139}

\definecolor{com_color}{RGB}{0,0,139}

\definecolor{orange_color}{RGB}{255,148,63}

\definecolor{gray_color}{RGB}{169,169,169}

\definecolor{lightgray}{RGB}{220,220,220}

\definecolor{lightgreen}{RGB}{179,207,176}

\definecolor{lightblue}{RGB}{181,209,230}

\usepackage[preprint, nonatbib]{nips_2023}
\usepackage{pdfpages}
\usepackage{times}
\usepackage{latexsym}
\usepackage{color}

\usepackage[T1]{fontenc}
\usepackage[utf8]{inputenc}

\usepackage{microtype}

\usepackage{inconsolata}
\usepackage{adjustbox}

\title{Harnessing GPT-4V(ision) for Insurance:\\ A Preliminary Exploration}

\author{Chenwei~Lin \\ Fudan University \\ \texttt{cwlin23@m.fudan.edu.cn} \And Hanjia~Lyu \\ University of Rochester \\ \texttt{hlyu5@ur.rochester.edu}  \AND Jiebo~Luo \\ University of Rochester \\ \texttt{jluo@cs.rochester.edu} \And Xian~Xu \\ Fudan University \\ \texttt{xianxu@fudan.edu.cn}}

\begin{document}
\maketitle
\begin{abstract}

The emergence of Large Multimodal Models (LMMs) marks a significant milestone in the development of artificial intelligence. 
Insurance, as a vast and complex discipline, involves a wide variety of data forms in its operational processes, including text, images, and videos, thereby giving rise to diverse multimodal tasks. 
Despite this, there has been limited systematic exploration of multimodal tasks specific to insurance, nor a thorough investigation into how LMMs can address these challenges. 
In this paper, we explore GPT-4V's capabilities in the insurance domain. We categorize multimodal tasks by focusing primarily on visual aspects based on types of insurance  (\eg auto, household/commercial property, health, and agricultural insurance) and insurance stages (\eg risk assessment, risk monitoring, and claims processing). 
Our experiment reveals that GPT-4V exhibits remarkable abilities in insurance-related tasks, demonstrating not only a robust understanding of multimodal content in the insurance domain but also a comprehensive knowledge of insurance scenarios. 
However, there are notable shortcomings: GPT-4V struggles with detailed risk rating and loss assessment, suffers from hallucination in image understanding, and shows variable support for different languages. 
Through this work, we aim to bridge the insurance domain with cutting-edge LMM technology, facilitate interdisciplinary exchange and development, and provide a foundation for the continued advancement and evolution of future research endeavors.

\end{abstract}

\tableofcontents
\newpage
\listoffigures
\newpage

\section{Introduction}\label{sec:intro}
\subsection{Motivation and Overview}
Traditionally, the insurance industry has been viewed as primarily focused on text and mathematical formulas~\cite{balona2023actuarygpt}. However, the reality is that insurance companies undertake a broad array of multimodal tasks that are crucial for the effective management of their operations~\cite{Kailan2018,xu2020framework}. 
These tasks span various \textit{types} of insurance and different \textit{stages} of the business cycle, including underwriting, loss assessment, and claims processing~\cite{smietanka2021algorithms,zarifis2023evaluating}. The integral role of these multimodal tasks is vital to the functioning of the contemporary insurance industry.
The evolution of insurance technology has witnessed the integration of AI technologies like computer vision,  significantly impacting the industry~\cite{sahni2020insurance,fernando2022automated,zhang2020automatic,li2018anti,hsu2020collision,dey2022survey,cao2020insurtech}.
However, these technologies often provide visual assistance for specific tasks and do not integrate well with multimodal tasks. 
Furthermore, these AI solutions often lack the specialized knowledge and contextual understanding essential for the insurance sector. As a result, the prevalent approach continues to prioritize human expertise, with AI serving as a supplementary resource~\cite{cao2020insurtech, balasubramanian2018insurance,stoeckli2018exploring}. 

The advent of Large Language Models (LLM) marks a significant milestone in the development of artificial intelligence. The debut of ChatGPT~\cite{chatgpt} in November 2022 showed its exceptional general capabilities across a wide range of text-based tasks~\cite{brown2020language, wei2022chain, ouyang2022training,lyu2023llm}. It also excels in the finance and insurance domains, showcasing its potential for transforming these industries~\cite{balona2023actuarygpt,chen2023disc, zhao2024revolutionizing,wu2023bloomberggpt}.
Building upon the foundation of text modeling, OpenAI took a step further by introducing Large Multimodal Models (LMM), specifically GPT-4V(ision)~\cite{openai2023gpt4}, which adds the ability to process visual inputs. 
GPT-4V has exhibited impressive capabilities in many general vision and language tasks~\cite{yang2023dawn,zhang2023gpt,wu2023early,zhang2024cocot}. Beyond the general applications, researchers have begun to focus on specialized domains, exploring the capability of GPT-4V in fields such as health~\cite{wu2023can,li2023comprehensive,ferber2024context}, agriculture~\cite{tan2023promises}, transportation~\cite{cui2024survey,wen2023road}, social media~\cite{lyu2023gpt,lyu2024human}, and biology~\cite{zheng2024exploring}. 
These studies not only validate GPT-4V's proficiency in specific professional domains but also provide valuable examples for other fields involved in multimodal learning.

The emergence of large multimodal models opens a new possibility for the insurance industry: Could the extensive general knowledge and robust multimodal understanding capabilities of large multimodal models address the diverse multimodal tasks in insurance scenarios?
Our research aims to explore the effectiveness of GPT-4V in multimodal tasks within the insurance domain, examining its capabilities and limitations. Our study is driven by the following questions:

\begin{itemize}[leftmargin=*]
\item \textbf{Can GPT-4V effectively handle multimodal tasks across different stages of insurance?} According to the insurance value chain theory~\cite{eling2018impact,eling2022impact}, the business stages of insurance include product design and development, sales and distribution, underwriting and pricing, claims processing, customer service, and asset management. Each stage incorporates various multimodal tasks to support its effective operation.  For instance, during the underwriting stage of health insurance, tasks necessitate engaging with multimodal data such as historical claims records, physical examination reports, and medical examination result images to perform an in-depth and detailed analysis and evaluation of the insured's risk status. Similarly, in the claims processing stage of auto insurance, tasks involve utilizing multimodal data like accident scene photographs and accident reports to aid in the precise determination of accident liability and damage assessment.

\item \textbf{Can GPT-4V effectively process multimodal tasks across different types of insurance?} Insurance spans across a wide array of categories, including auto insurance, property insurance, health insurance, and others. While the core principles of insurance remain consistent, the knowledge systems and business models of different types vary significantly. Even within the same business stage, the multimodal tasks for different types of insurance are not identical. For instance, in the underwriting stage, property insurance focuses on the risk status of the property, while health insurance focuses on the health condition of the individual. This requires the model to possess rich insurance knowledge and the ability to understand various scenarios.
\end{itemize}

\begin{figure}[ht]
\centering
\includegraphics[scale=0.75]{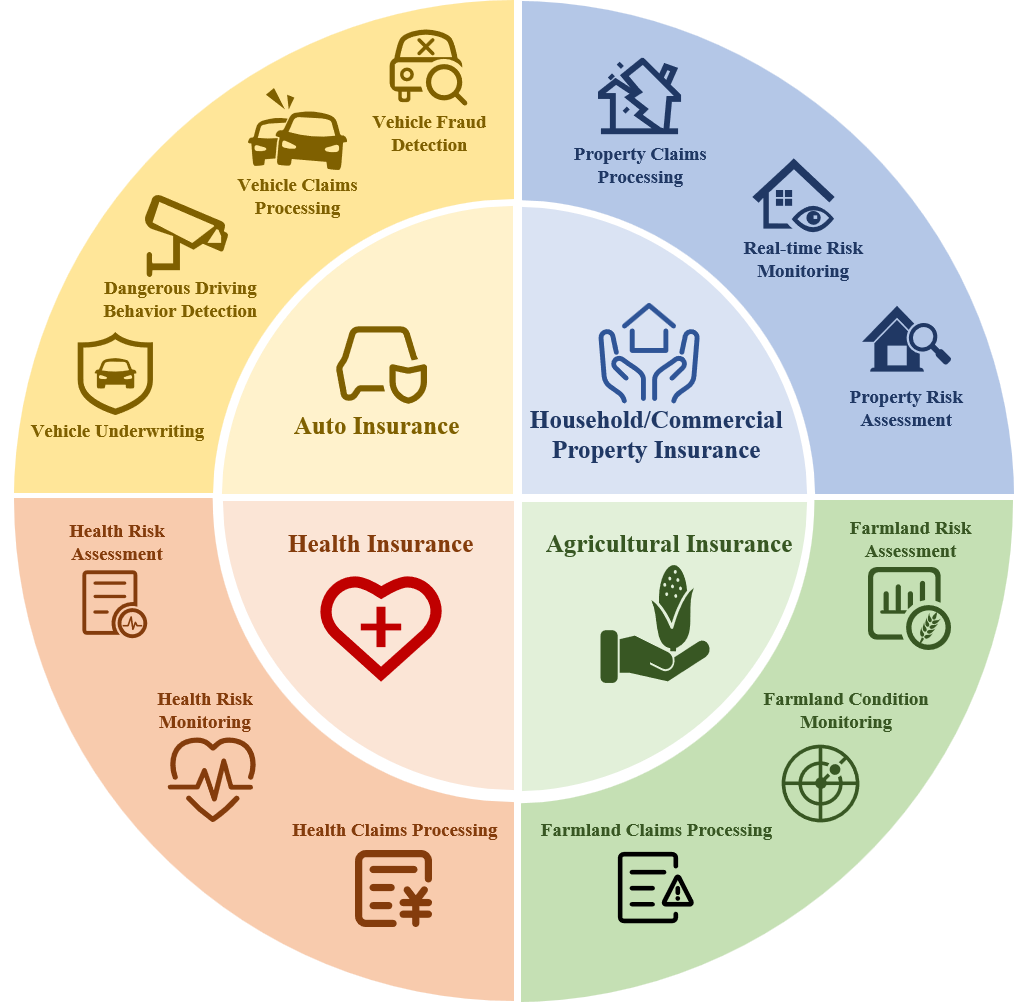}
\caption[Section~\ref{sec:intro}: overview of the task classification in the insurance domain]{An overview of the task classification in the insurance domain.}
\label{task}
\end{figure}

\begin{figure}[ht] 
  \centering 
  \begin{subfigure}[b]{0.95\textwidth} 
  \centering 
    \includegraphics[width=0.95\textwidth]{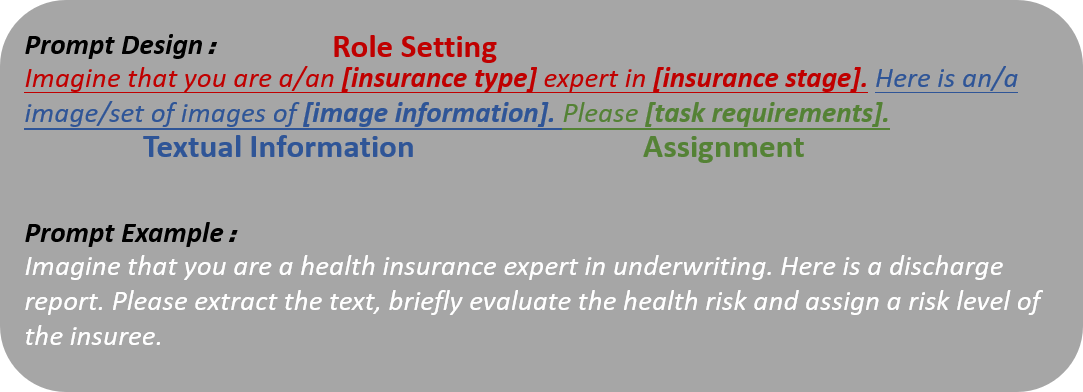} 
    \caption[Section~\ref{sec:intro}: prompt design of our research]{The prompt design of our research.}
    \label{prompt} 
  \end{subfigure}
  \hfill 
  \begin{subfigure}[b]{0.95\textwidth} 
  \centering 
    \includegraphics[width=0.95\textwidth]{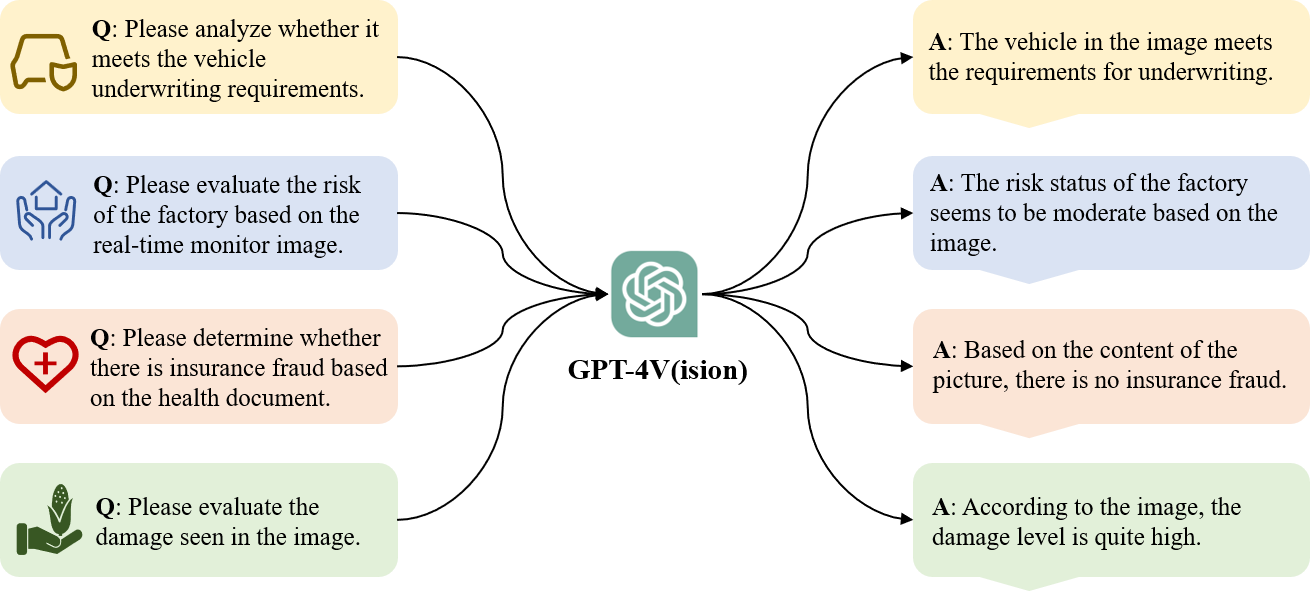}
    \caption[Section~\ref{sec:intro}: pipeline of GPT-4V for insurance tasks]{The pipeline of GPT-4V for insurance tasks.}
    \label{pipeline}
  \end{subfigure}
  \caption[Section~\ref{sec:intro}: overview of our approach in exploring GPT-4V(ision)]{An overview of our approach in exploring GPT-4V(ision).} 
  \label{intergrate}
\end{figure}

\subsection{Our Approach in Exploring GPT-4V(ision)}

Previous research has seen extensive qualitative and quantitative analysis by numerous scholars to investigate the capabilities of GPT-4V. However, executing quantitative assessments in the insurance sector is particularly challenging compared to other fields. This challenge arises primarily due to the absence of a systematic framework for categorizing and classifying computer vision tasks specific to the insurance industry, coupled with a lack of standardized benchmark datasets. Consequently, this study aims to define multimodal tasks within the insurance domain and focuses on conducting qualitative tests through various case studies.

\paragraph{Task Classification.} In identifying and classifying tasks, we focus on two dimensions: the types and stages of insurance as shown in Figure~\ref{task}. Insurance can be classified into two main types based on the insurance subject matter: personal insurance and general insurance (also known as property and casualty insurance)~\cite{weedige2019decision, driver2018insurance, murzalieva2023state}. Personal insurance covers individuals' life, health, and physical well-being, including life insurance and health insurance. General insurance, on the other hand, covers property and third-party liabilities,  including property insurance, auto insurance, and others. To ensure that our research encompasses a comprehensive and representative view of the insurance industry, we have selected typical insurance types from both categories: health insurance representing personal insurance, while auto insurance, household/commercial property insurance, and agricultural insurance representing general insurance. These insurance types differ in their objects of coverage and modes of operation. Building on this, we further classify tasks from the perspective of insurance stages, identifying three core tasks: risk assessment, risk monitoring, and claims processing. Each insurance type encompasses these stages, and each stage is present across the various insurance types. 

\paragraph{Testing Pipeline and Prompt Design.}
We assess GPT-4V using its online chat interface.\footnote{\url{https://chat.openai.com/}} Our evaluation methodology involves crafting distinct prompt statements tailored to various task categories within the insurance sector. Each prompt incorporates three essential elements: role setting, textual information, and a specific task. These prompts are customized to align with the particular type of insurance, the stage of the insurance process, the inclusion of image information, and the specific task requirements, as depicted in Figure~\ref{prompt}. The detailed testing pipeline is illustrated in Figure~\ref{pipeline}.

In Section~\ref{sec:experiment}, we present our exploration of GPT-4V across four insurance types: auto insurance (Section~\ref{sec:auto}), household/commercial property insurance (Section~\ref{sec:property}), health insurance (Section~\ref{sec:health}) and agriculture insurance (Section~\ref{sec:agri}). In Section~\ref{sec:Challenges and Opportunities}, we
discuss the challenges encountered and outline potential future opportunities. Our research is concluded in Section~\ref{sec:Conclusions}.

\paragraph{Sample Selection.} 
To accurately represent the multimodal task demands in insurance scenarios, while also considering the complexity and diversity of insurance contexts, we have created our test sample set using publicly available online images and video screenshots.

\paragraph{Ethical Consideration.} In conducting the experiments, we pay rigorous attention to ethical considerations, particularly in relation to privacy and data protection. The cases used for testing GPT-4V in the insurance domain inherently contain sensitive information, including vehicle identification numbers (VINs), medical records, and personal information among other private data. To adhere to the common ethical guidelines and to ensure compliance with data protection regulations, all such sensitive information is subjected to anonymization and privacy-enhancing processes prior to analysis. The anonymization process is designed to be irreversible, ensuring that there is no feasible method to trace back or reconstruct the original data from the information used in our experiments. The insights and conclusions outlined in this research are based on the data available and the capabilities of the GPT-4V model as of April 2024. This work seeks to contribute to the continuous dialogue within the insurance domain and AI research. It is crucial to acknowledge that the findings may not fully capture all the nuances and complexities inherent in practical insurance scenarios.

\newpage

\section{Experiments}\label{sec:experiment}
\subsection{Auto Insurance}\label{sec:auto}
Auto insurance is intended to offer policyholders financial protection against losses resulting from incidents such as accidents or theft. It primarily covers liabilities in property damage, legal responsibility, and medical expenses~\cite{naic}. According to projections by SkyQuest~\cite{skyquest}, the global market size of auto insurance is expected to rise from \$910.9 billion in 2023 to \$1610.13 billion by 2031, making it one of the most crucial types of insurance due to the ubiquity of automobiles and the mandatory nature of auto insurance. With the market's rapid growth, reliance on manual processes for on-site underwriting or claims handling can no longer meet the market's demand~\cite{patil2017deep,singh2019automating}. The importance of insurtech~\cite{cao2020insurtech}, represented by the computer vision technology, is increasingly highlighted, with more insurance companies optimizing business processes through online photo review~\cite{li2018anti,hsu2020collision,li2007applying,dhieb2019very,kyu2020car,yang2023auto}. For example, policyholders can apply for claims by uploading images of the accident scene along with related documents, while insurance companies use computer vision technologies to automate the analysis of car damage, significantly enhancing operational efficiency. A recent study~\cite{yang2023dawn} has explored GPT-4V's abilities in damage assessment and report generation within auto insurance. Building on this foundation, we delve deeper, categorizing multimodal tasks in the auto insurance domain into (1) vehicle underwriting assessment, (2) dangerous driving behavior detection, (3) vehicle damage assessment, and (4) insurance fraud detection. Moreover, we conduct exploratory tests for each task.

\subsubsection{Vehicle Underwriting}\label{sec:vehicle underwriting}
Auto insurance underwriting is the process of evaluating and deciding whether to underwrite a vehicle insurance policy. This involves reviewing key risk factors including age, gender, marital status, car model, car condition, mileage, VIN, \textit{etc}. to determine the level of risk and support the pricing of the policy~\cite{conrad2009underwriting,fan2017comparison,acharya2019mileage}. In their day-to-day operations, insurance companies increasingly utilize computer vision technologies to automate the extraction and analysis of these risk factors, thereby enhancing operational efficiency~\cite{acharya2019mileage, fang2016fine}.

\begin{figure}
\centering 
\includegraphics[width=1\textwidth,height=1.5\textwidth]{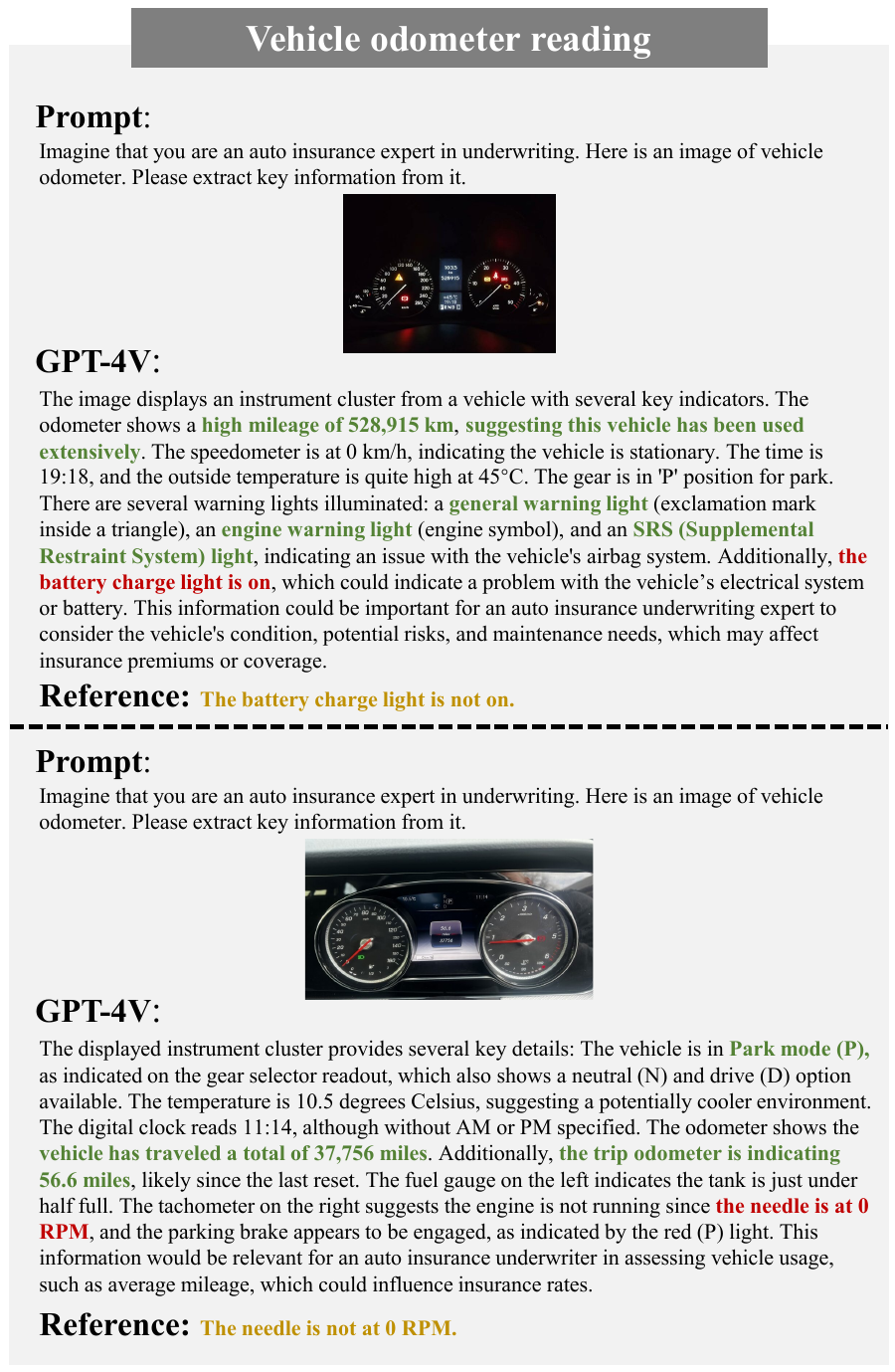} 
\caption[Section~\ref{sec:vehicle underwriting}: vehicle odometer reading test case]{The test case of GPT-4V's capability in vehicle odometer reading. The correct answer parts, incorrect answer parts, and references are highlighted in \textcolor[RGB]{84,130,53}{\textbf{green}}, \textcolor[RGB]{192,0,0}{\textbf{red}} and \textcolor[RGB]{191,144,0}{\textbf{yellow}}, respectively.} 
\label{odometer reading}
\end{figure}

We design two distinct tasks to evaluate GPT-4V's capabilities in vehicle underwriting: the first task involves extracting key information from an image of a vehicle's odometer. Our prompt is designed as follows: “\textit{Imagine that you are an auto insurance expert in underwriting. Here is an image of a vehicle odometer. Please extract key information from it.}” The experimental results (see Figure~\ref{odometer reading}) indicate that GPT-4V can accurately extract key information from the odometer (\eg “\textit{... high mileage of 528,915 km ...}”, “\textit{... There are several warning lights illuminated ...}”), although there are minor mistakes (\eg “\textit{... the battery charge light is on ...}”, “\textit{... the needle is at 0 RPM ...}”).

\begin{figure}
\centering 
\includegraphics[width=1\textwidth,height=1.48\textwidth]{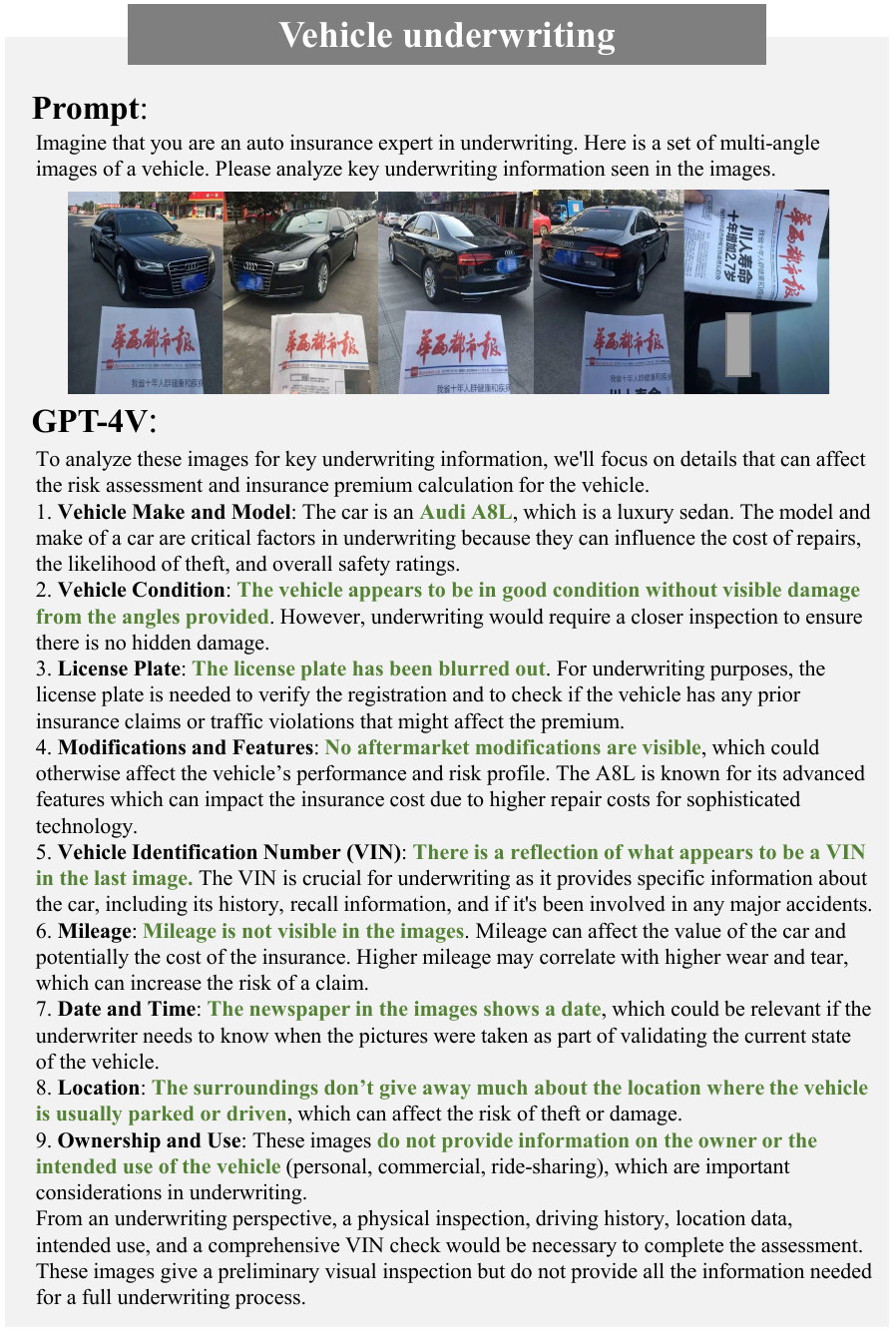} 
\caption[Section~\ref{sec:vehicle underwriting}: vehicle underwriting test case]{The test case of GPT-4V's capability in vehicle underwriting. The relevant parts are highlighted in \textcolor[RGB]{84,130,53}{\textbf{green}}. }
\label{vehicle underwriting} 
\end{figure}

The second task involves conducting a comprehensive analysis of a vehicle's model and condition based solely on its exterior appearance from multiple angles. Our prompt is: “\textit{Imagine that you are an auto insurance expert in underwriting. Here is a set of multi-angle images of a vehicle. Please analyze key underwriting information seen in the images.}” The experimental results (see Figure~\ref{vehicle underwriting}) demonstrate that GPT-4V possesses strong analytical capabilities for vehicle underwriting. It can analyze and extract necessary information for underwriting from various dimensions, including both present (\eg vehicle model, vehicle condition) and absent information (\eg license, mileage). Moreover, an interesting finding is GPT-4V's ability to comprehend the role of a newspaper in the images as an indication of the date of photography.

\subsubsection{Dangerous Driving Behavior Detection}\label{sec: dangerous driving behavior detection}
Dangerous driving behavior detection refers to identifying and issuing timely warnings for any anomalies, such as drowsy driving, making calls, driving against traffic, and speeding~\cite{ma2017drivingsense,wei2023lightweight}. Dangerous driving behaviors are among the primary causes of traffic accidents~\cite{kaplan2015driver}. To mitigate the probability and scale of losses, insurance companies utilize a variety of data sources (\eg OBD devices, smartphones, cameras) for real-time monitoring of driving behaviors~\cite{arumugam2019survey}. It is a more direct and effective way to collect images through camera and use computer vision technologies to detect dangerous driving behaviours, because of the richness of the image information and the advancement of computer vision~\cite{li2016dangerous,lashkov2019driver}. 

\begin{figure}
\centering 
\includegraphics[width=1\textwidth,height=1.4\textwidth]{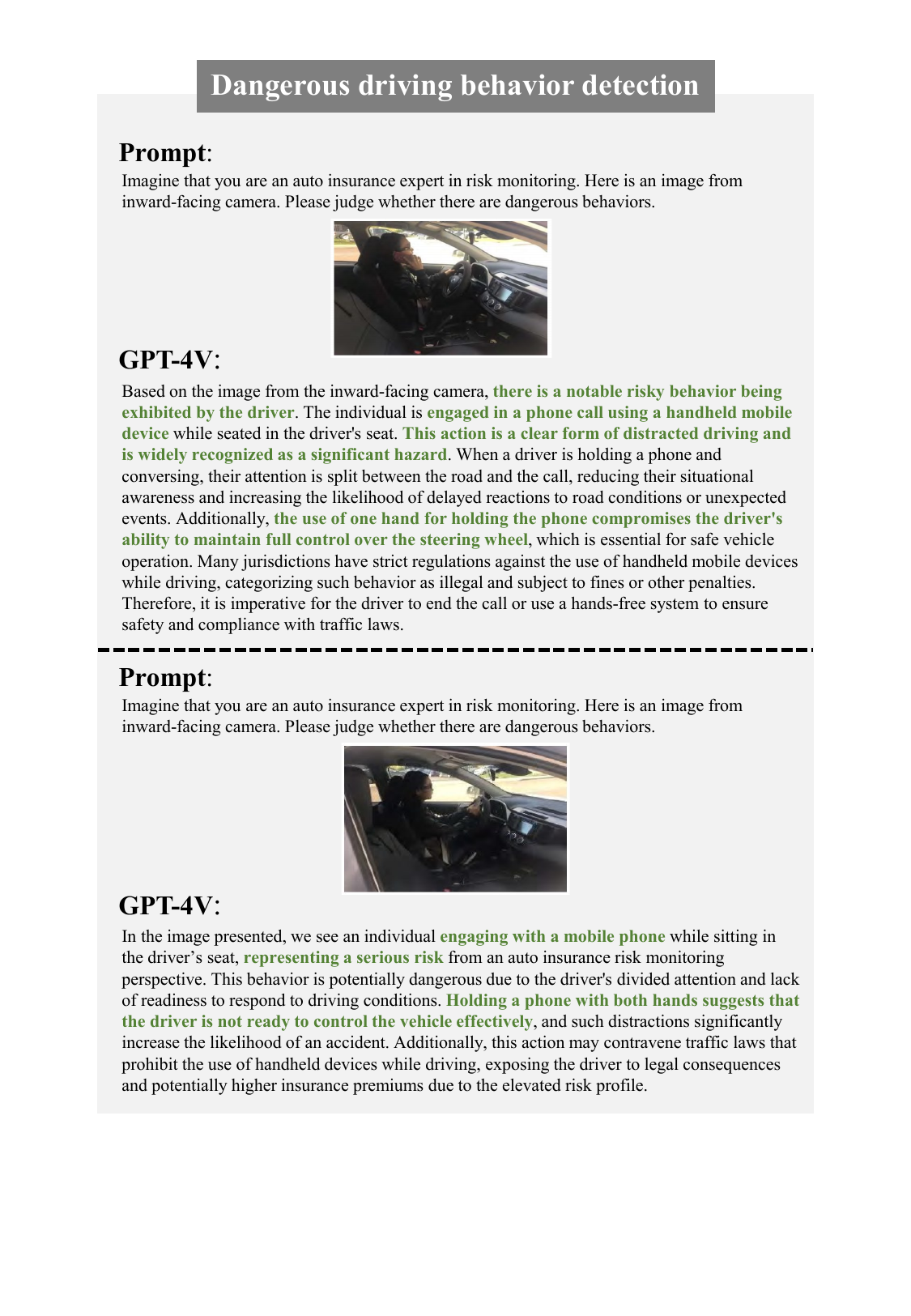} 
\caption[Section~\ref{sec: dangerous driving behavior detection}: dangerous driving behavior detection test case 1]{The test case of GPT-4V's capability in dangerous driving behavior detection through inward-facing camera. The relevant parts are highlighted in \textcolor[RGB]{84,130,53}{\textbf{green}}.} 
\label{inward-facing camera} 
\end{figure}

We design two distinct tasks to test the capabilities of GPT-4V in detecting dangerous driving behaviors: the first task involves monitoring the driver's behavior through an inward-facing camera. The prompt is: “\textit{Imagine that you are an auto insurance expert in risk monitoring. Here is an image from an inward-facing camera. Please judge whether there are dangerous behaviors.}” The results (see Figure~\ref{inward-facing camera}) indicate that GPT-4V can effectively identify the driver's dangerous behaviors within the vehicle (\eg “\textit{... phone call ...}”, “\textit{... engaging with a mobile phone ...}”).

\begin{figure}
\centering 
\includegraphics[width=1\textwidth,height=1.52\textwidth]{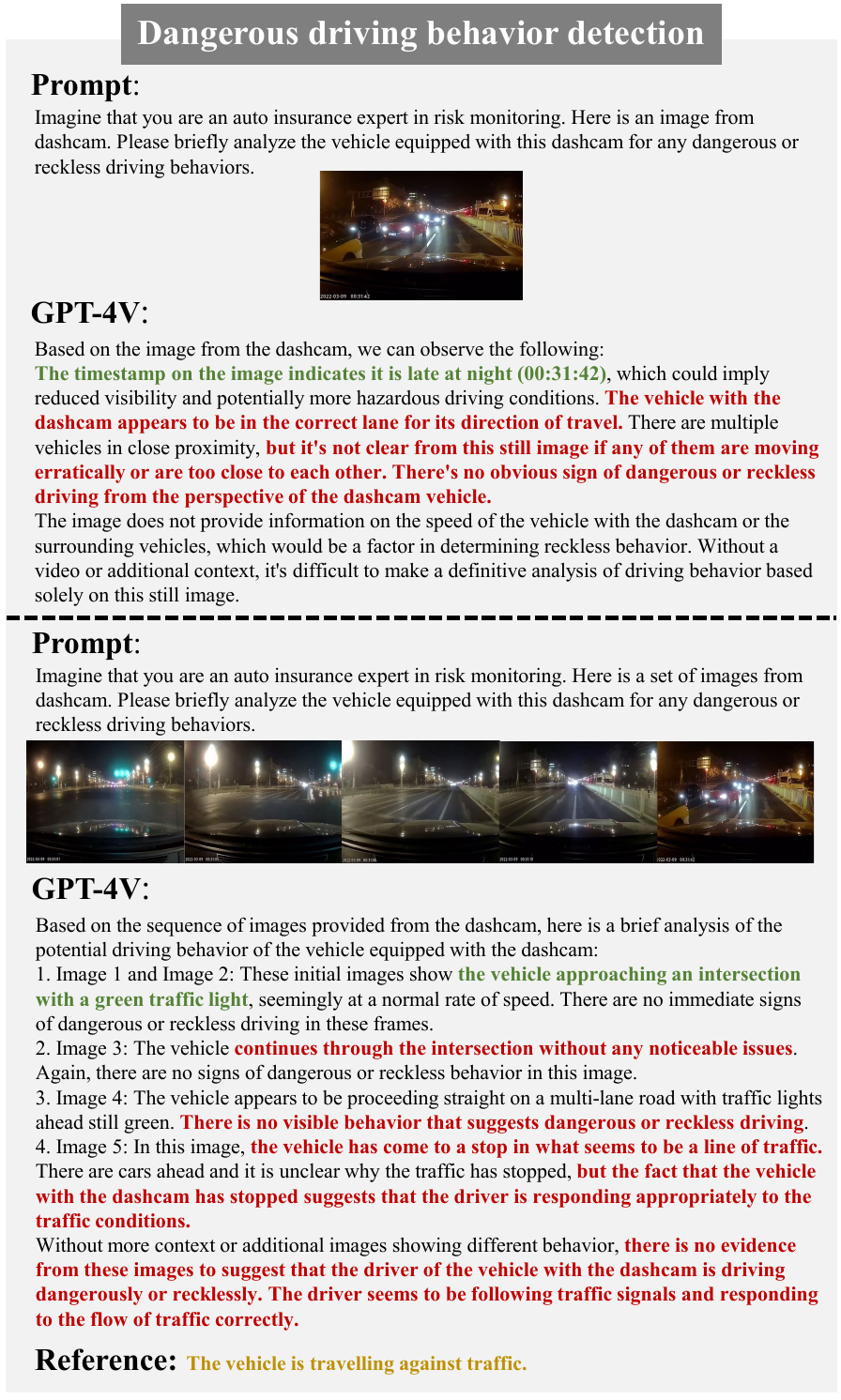} 
\caption[Section~\ref{sec: dangerous driving behavior detection}: dangerous driving behavior detection test case 2]{The test case of GPT-4V's capability in dangerous driving behavior detection through dashcam. The correct answer parts, incorrect answer parts, and references are highlighted in \textcolor[RGB]{84,130,53}{\textbf{green}}, \textcolor[RGB]{192,0,0}{\textbf{red}} and \textcolor[RGB]{191,144,0}{\textbf{yellow}}, respectively.} 
\label{dashcam} 
\end{figure}

The second task involves monitoring the driver's behavior through a dashcam. In order to provide a comprehensive demonstration of the driving process, we use both single image inputs and second-by-second captures of multiple image inputs for the same case. The prompt is: “\textit{Imagine that you are an auto insurance expert in risk monitoring. Here is an image / a set of images from a dashcam. Please briefly analyze the vehicle equipped with this dashcam for any dangerous or reckless driving behaviors.}” The findings (see Figure~\ref{dashcam}) reveal that GPT-4V can not identify instances of driving against the direction of traffic, whether in individual images or in images captured sequentially.

\subsubsection{Vehicle Claims Processing}\label{sec: vehicle damage evaluation}
Vehicle claims processing involves a series of standardised operations: checking the make and model of the vehicle, locating and classifying the type of damage, and determining the extent of the damage~\cite{singh2019automating}. In the past, claims processing was a collaborative process involving a number of parties, including insurance adjusters and automotive repair technicians. Not only was it time-consuming, it was also prone to human error, resulting in inaccurate damage assessments~\cite{kannan2023cda}. According to a research report by Verisk~\cite{verisk}, the U.S. auto insurance market loses around \$29 billion annually due to human error and information omissions in the vehicle damage detection and assessment process. As a result, a growing number of insurance companies are beginning to incorporate computer vision technologies into claims processing to improve the accuracy and timeliness of assessments~\cite{zhang2020automatic,poon2021modeling,doshi2023vehicle}.

\begin{figure}
\centering 
\includegraphics[width=1\textwidth,height=1.4\textwidth]{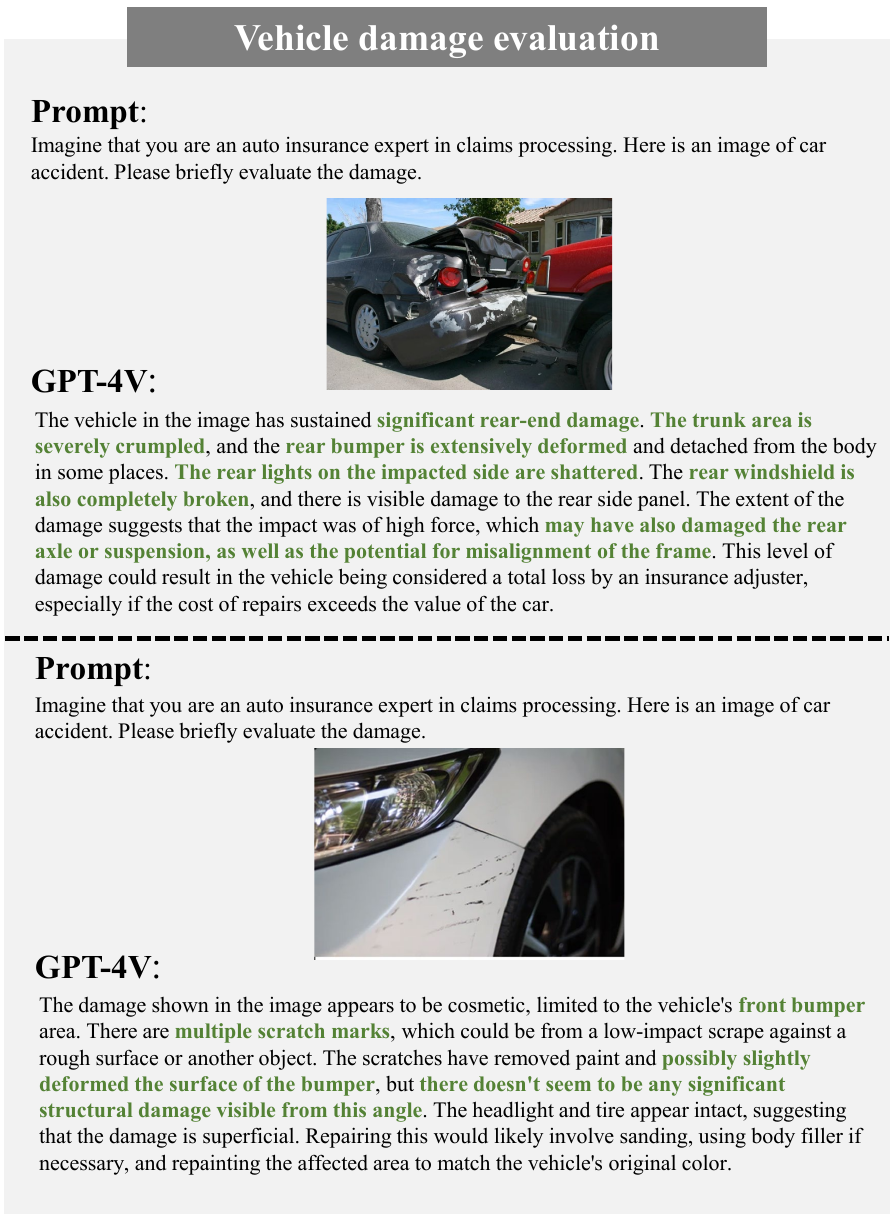} 
\caption[Section~\ref{sec: vehicle damage evaluation}: vehicle damage evaluation test case]{The test case of GPT-4V's capability in vehicle damage evaluation. The relevant parts are highlighted in \textcolor[RGB]{84,130,53}{\textbf{green}}.} 
\label{vehicle damage evaluation} 
\end{figure}

We design two tasks to test the capabilities of GPT-4V in vehicle claims processing: the first task involves a brief evaluation of the vehicle's damage condition, where we ask GPT-4V to summarize the overall damage to the vehicle. The prompt is: “\textit{Imagine that you are an auto insurance expert in claims processing. Here is an image of a car accident. Please briefly evaluate the damage.}” The results (see Figure~\ref{vehicle damage evaluation}) indicate that GPT-4V can accurately determine the extent and location of damage to the vehicle (\eg “\textit{... the rear bumper is extensively deformed ...}”, “\textit{... slightly deformed the surface ...}”).

\begin{figure}
\centering 
\includegraphics[width=1\textwidth,height=1.5\textwidth]{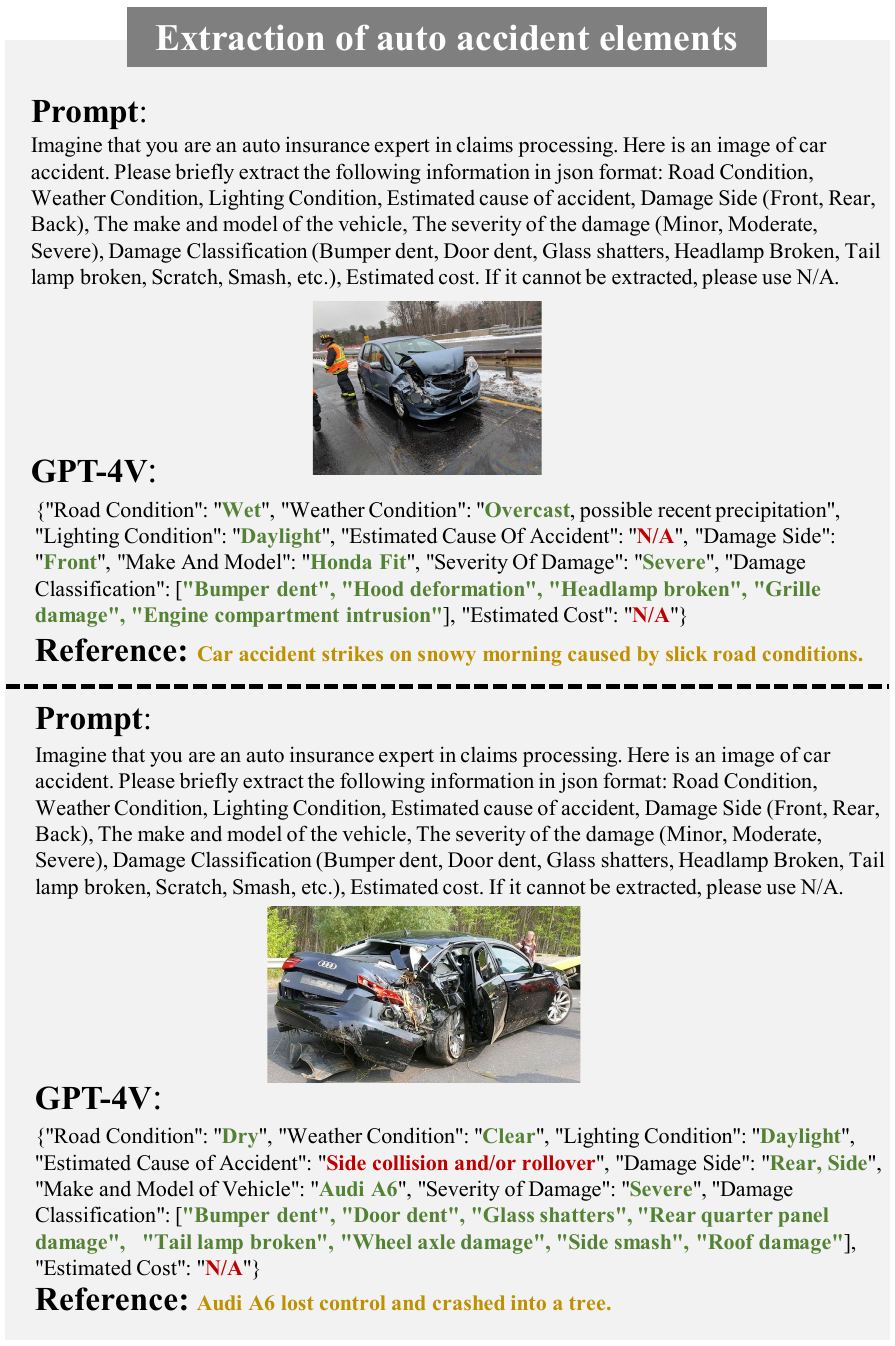} 
\caption[Section~\ref{sec: vehicle damage evaluation}: extraction of auto accident elements test case]{The test case of GPT-4V's capability in extraction of auto accident elements. The correct answer parts, incorrect answer parts, and references are highlighted in \textcolor[RGB]{84,130,53}{\textbf{green}}, \textcolor[RGB]{192,0,0}{\textbf{red}} and \textcolor[RGB]{191,144,0}{\textbf{yellow}}, respectively.} 
\label{extraction of auto accident elements test case} 
\end{figure}

The second task requires the extraction of structured information about the vehicle incident to assist in the damage report generation. The prompt is: “\textit{Imagine that you are an auto insurance expert in claims processing. Here is an image of a car accident. Please extract the following information in JSON format.}” According to the results (Figure~\ref{extraction of auto accident elements test case}), GPT-4V effectively identifies environmental conditions such as road, weather, lighting conditions, and vehicle information including make and model, and assesses the extent and location of the damage. However, it performs poorly in analyzing the cause of the accident and in assessing the cost of the damage.

\subsubsection{Fraud Detection}\label{sec: fraud detection}
Auto insurance claims processing is characterized by its high frequency, low value, and automated nature, making it challenging for insurance companies to effectively detect fraud~\cite{li2018anti}. It is reported that a significant proportion of auto insurance claims, ranging from 21\% to 36\%, are believed to include fraudulent elements. However, only a small fraction of these suspected cases, less than 3\%, result in prosecution~\cite{nian2016auto}. Various computer vision technologies have been proposed to combat auto insurance fraud~\cite{yang2023auto, maiano2023deep}. These technologies analyze images from accident scenes to identify potential fraudulent activities, such as fabricating non-existent damages or exaggerating the amount of actual losses~\cite{benedek2022automobile}.

\begin{figure}
\centering 
\includegraphics[width=1\textwidth,height=1.5\textwidth]{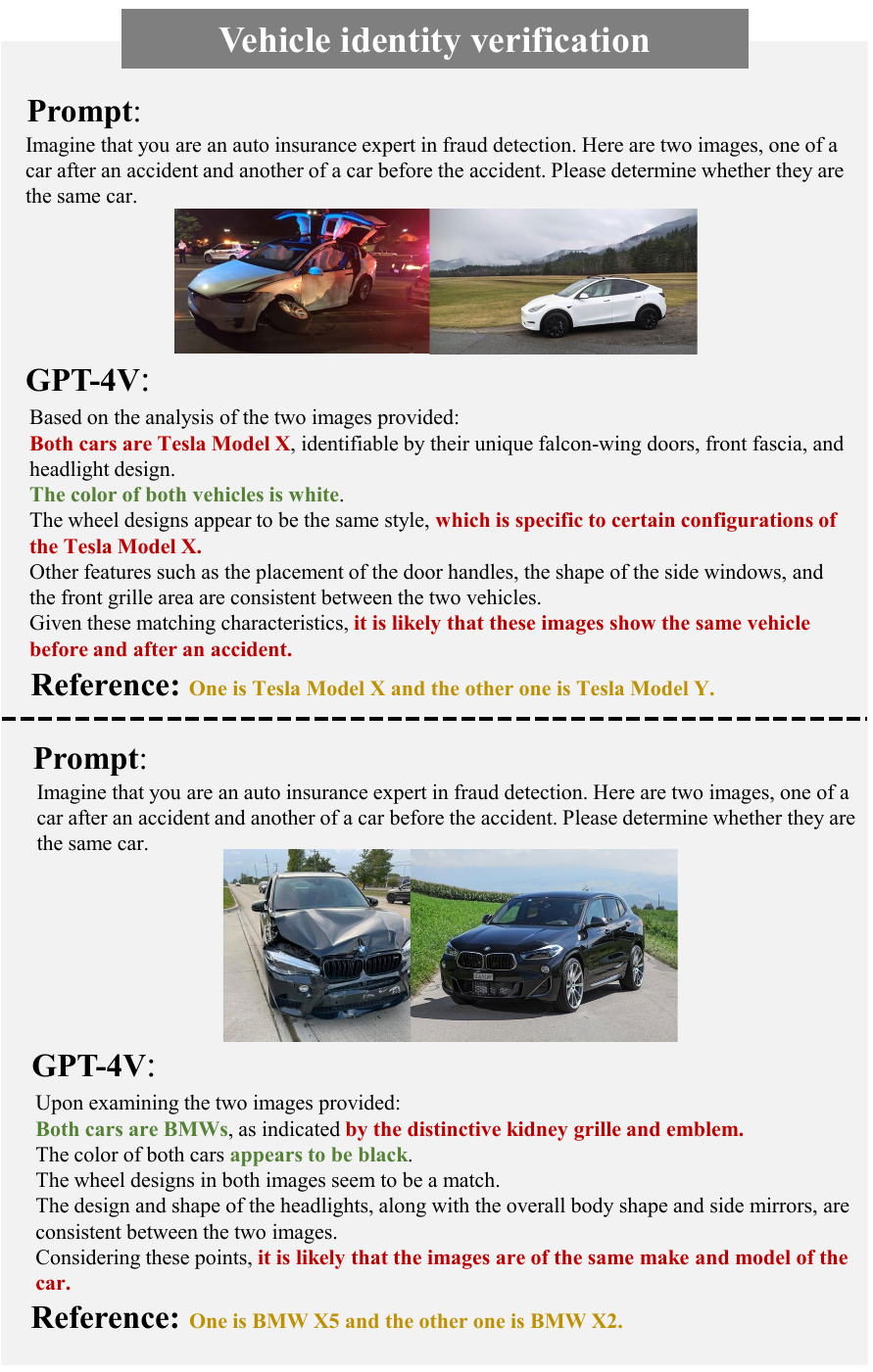} 
\caption[Section~\ref{sec: fraud detection}: vehicle identity verification test case]{The test case of GPT-4V's capability in vehicle identity verification. The correct answer parts, incorrect answer parts, and references are highlighted in \textcolor[RGB]{84,130,53}{\textbf{green}}, \textcolor[RGB]{192,0,0}{\textbf{red}} and \textcolor[RGB]{191,144,0}{\textbf{yellow}}, respectively.}
\label{vehicle identify verification} 
\end{figure}

To assess the fraud detection capabilities of GPT-4V, we design two tasks. The first task involves comparing user-uploaded post-accident images of a vehicle with pre-accident images of the same vehicle to assess if they match. For this purpose, we intentionally select images of similar-looking but different models from the same brand, such as Tesla Model X and Model Y, BMW X5 and BMW X2. The prompt provided is: “\textit{Imagine that you are an auto insurance expert in fraud detection. Here are two images, one of a car after an accident and another of a car before the accident. Please determine whether they are the same car.}” The test results (see Figure~\ref{vehicle identify verification}) reveal that GPT-4V struggles to distinguish between different models and incorrectly identifies them as the same car.

\begin{figure}
\centering 
\includegraphics[width=1\textwidth,height=1.5\textwidth]{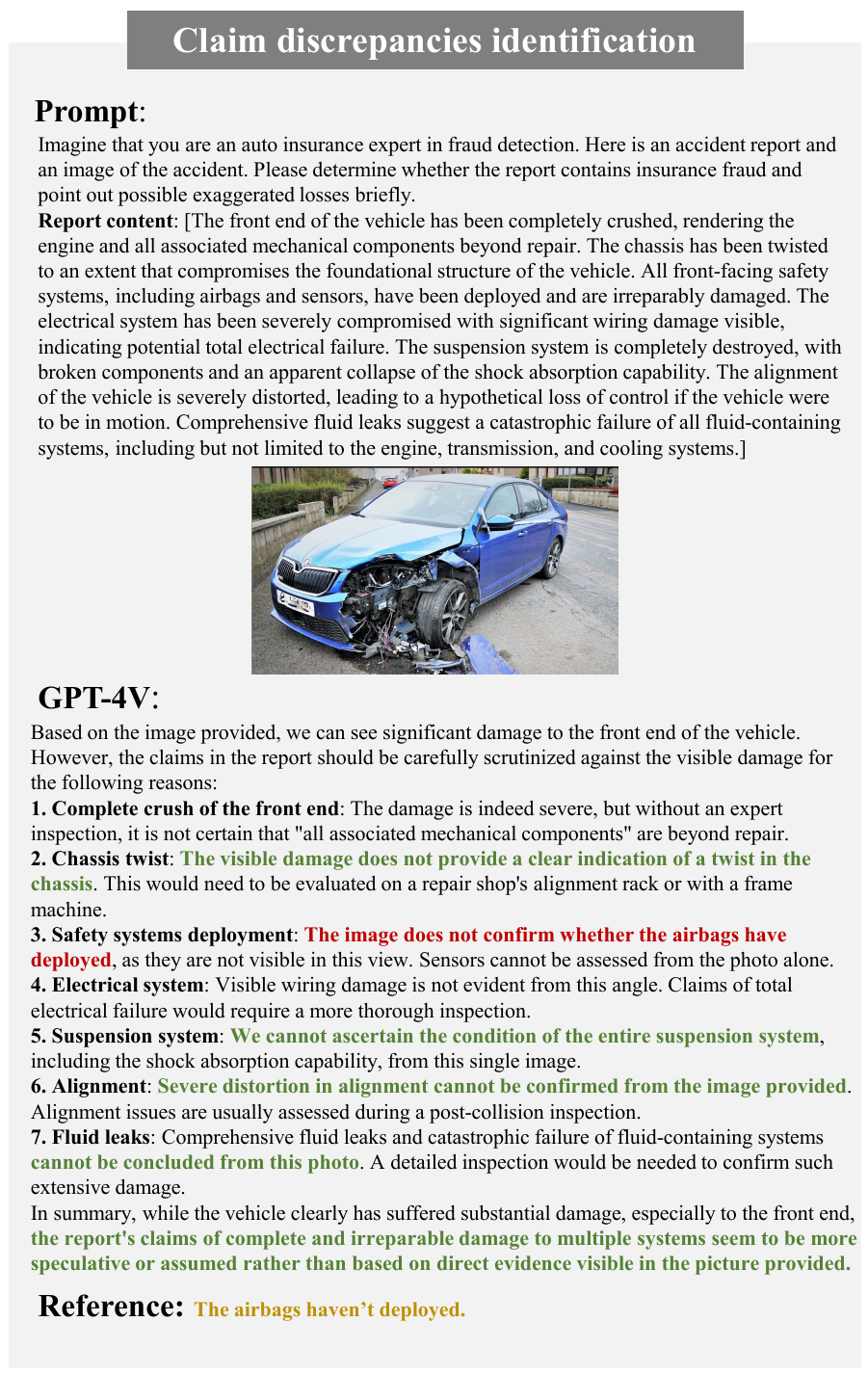} 
\caption[Section~\ref{sec: fraud detection}: claim discrepancies identification test case]{The test case of GPT-4V's capability in claim discrepancies identification. The correct answer parts, incorrect answer parts, and references are highlighted in \textcolor[RGB]{84,130,53}{\textbf{green}}, \textcolor[RGB]{192,0,0}{\textbf{red}} and \textcolor[RGB]{191,144,0}{\textbf{yellow}}, respectively.}
\label{claim discrepencies identification} 
\end{figure}

The second task involves comparing vehicle accident images and damage reports to determine if there are any inconsistencies or exaggerations. The prompt is: “\textit{Imagine that you are an auto insurance expert in fraud detection. Here is an accident report and an image of the accident. Please determine whether the report contains insurance fraud and point out possible exaggerated losses briefly.}” The test results (see Figure~\ref{claim discrepencies identification}) show that GPT-4V questions certain details about the reported losses (\eg “\textit{... does not provide a clear indication of a twist in the chassis ...}”) and considers the contents of the report to be highly speculative or assumed.

\newpage

\subsection{Household and Commercial Property Insurance}\label{sec:property}

Household and commercial property insurance plays a crucial role not only in protecting individual and corporate assets but also in mitigating the potential negative impacts of various risks on economic activities~\cite{felsenstein2018household,hanson2020commercial}. With the advent of computer vision, household and commercial property insurance can benefit greatly~\cite{Kailan2018}. For instance, in the field of household property insurance, images of residential roofs can lead to a clearer understanding of roof conditions and help identify potential hazards~\cite{wang2018residential}. Likewise, in commercial property insurance, surveillance images can be instrumental in recognizing and alerting against dangerous behaviors among workers~\cite{fang2020computer}. In our study, we combine household and commercial property insurance for a collective analysis and evaluate the capabilities of GPT-4V from three perspectives: risk assessment, risk monitoring, and claims processing.

\subsubsection{Household and Commercial Property Risk Assessment}\label{sec: Household/Commercial Property Risk Assessment}

Risk assessment in household/commercial property insurance involves analyzing and evaluating the internal and external risk conditions of a building to assist in accurately pricing insurance policies~\cite{parvin2005risk}. Advancements in computer vision now enable its application in assessing both building structure and work safety risks~\cite{wang2018residential,rivera2020work}, providing effective tools for insurance companies. For example, CAPE Analytics\footnote{\url{https://capeanalytics.com/home-insurance-property-intelligence/}} uses intelligent algorithms to conduct precise risk analysis on the structural integrity and roofing of residential homes.

We design two tasks to evaluate the risk monitoring capabilities of GPT-4V: the first task involves evaluating the risk condition of residential house roofs. The prompt used is: “\textit{Imagine that you are a household property expert in risk assessment. Here is an image of a roof. Please evaluate the roof's condition and risk rank.}” The experimental results (see Figure~\ref{household property risk assessment}) indicate that GPT-4V is capable of accurately analyzing roofs in various conditions (\eg “\textit{... Roof shingles seem intact with no visible damage or algae ...}”, “\textit{... Missing shingles, curling, and warping shingles ...}”) and providing a reasonable risk rank.

\begin{figure}
\centering 
\includegraphics[width=1\textwidth,height=1.5\textwidth]{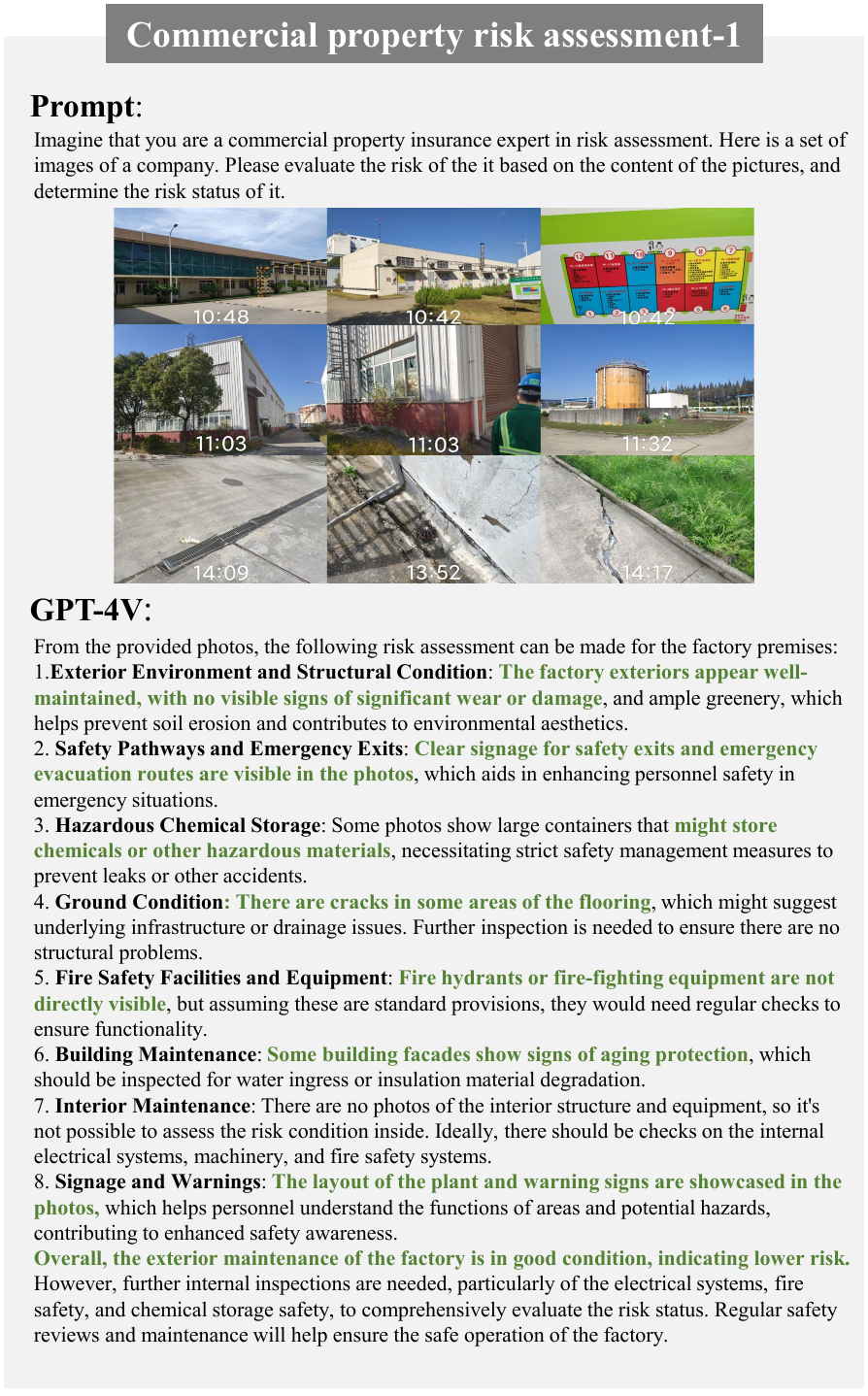} 
\caption[Section~\ref{sec: Household/Commercial Property Risk Assessment}: commercial property risk assessment test case 1]{The test case of GPT-4V's capability in commercial property risk assessment. The relevant parts are highlighted in \textcolor[RGB]{84,130,53}{\textbf{green}}.} 
\label{commercial property risk assessment-1} 
\end{figure}

\begin{figure}
\centering 
\includegraphics[width=1\textwidth,height=1.5\textwidth]{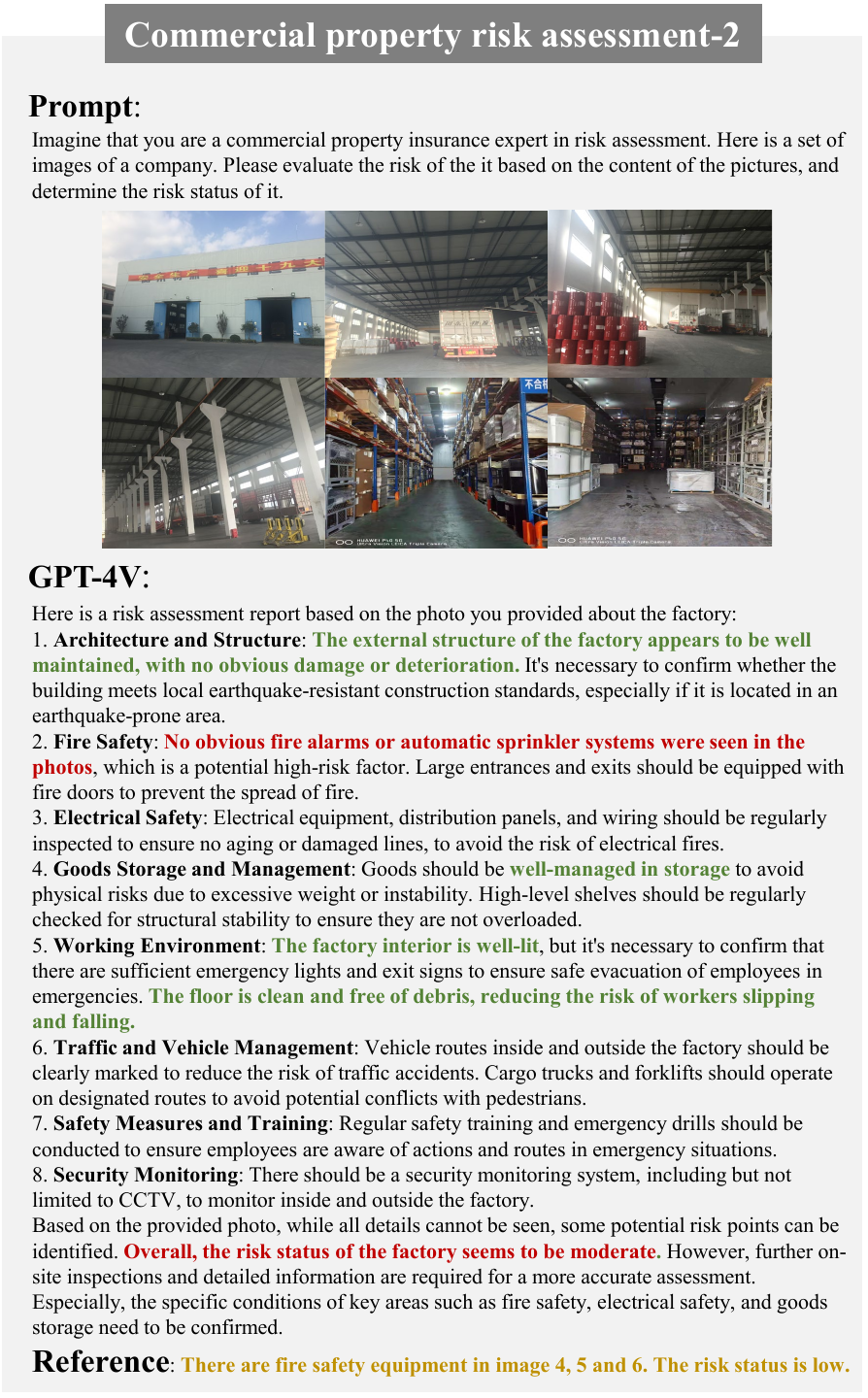} 
\caption[Section~\ref{sec: Household/Commercial Property Risk Assessment}: commercial property risk assessment test case 2]{The test case of GPT-4V's capability in commercial property risk assessment. The correct answer parts, incorrect answer parts, and references are highlighted in \textcolor[RGB]{84,130,53}{\textbf{green}}, \textcolor[RGB]{192,0,0}{\textbf{red}} and \textcolor[RGB]{191,144,0}{\textbf{yellow}}, respectively.}
\label{commercial property risk assessment-2} 
\end{figure}

The second task involves conducting a thorough risk assessment of a commercial facility. We use two sets of comprehensive, multi-angle photos of a company as input, with the following prompt: “\textit{Imagine that you are a commercial property insurance expert in risk assessment. Here is a set of images of a company. Please evaluate the risk of the it based on the content of the pictures, and determine the risk status of it.}" The experimental results (see Figures~\ref{commercial property risk assessment-1} and \ref{commercial property risk assessment-2}) suggest that GPT-4V can analyze and evaluate the risk condition from multiple dimensions such as fire safety, architecture, and structure. However, it still exhibits certain limitations in recognizing image details (\eg “\textit{... No obvious fire alarms or automatic sprinkler systems ...}”), which may lead to misjudgments in assessing risk conditions.

\subsubsection{Household and Commercial Property Real-time Risk Monitoring}\label{sec: Household/Commercial Property Real-time Risk Monitoring}

Risk monitoring in household and commercial property insurance involves identifying and detecting potential scenarios through surveillance footage that could harm the property~\cite{lopez2018review}. Continuous risk monitoring plays a crucial role in reducing risks and minimizing losses. Monitoring systems are commonly used in many homes and businesses. Insurance companies, by utilizing computer vision technologies, can timely capture environmental conditions, human behavior, and conditions of insured items, achieving continuous risk detection and safety alerts~\cite{dukuzumuremyi2014novel,pincott2022indoor,lee2022computer}.

\begin{figure}
\centering 
\includegraphics[width=1\textwidth,height=1.45\textwidth]{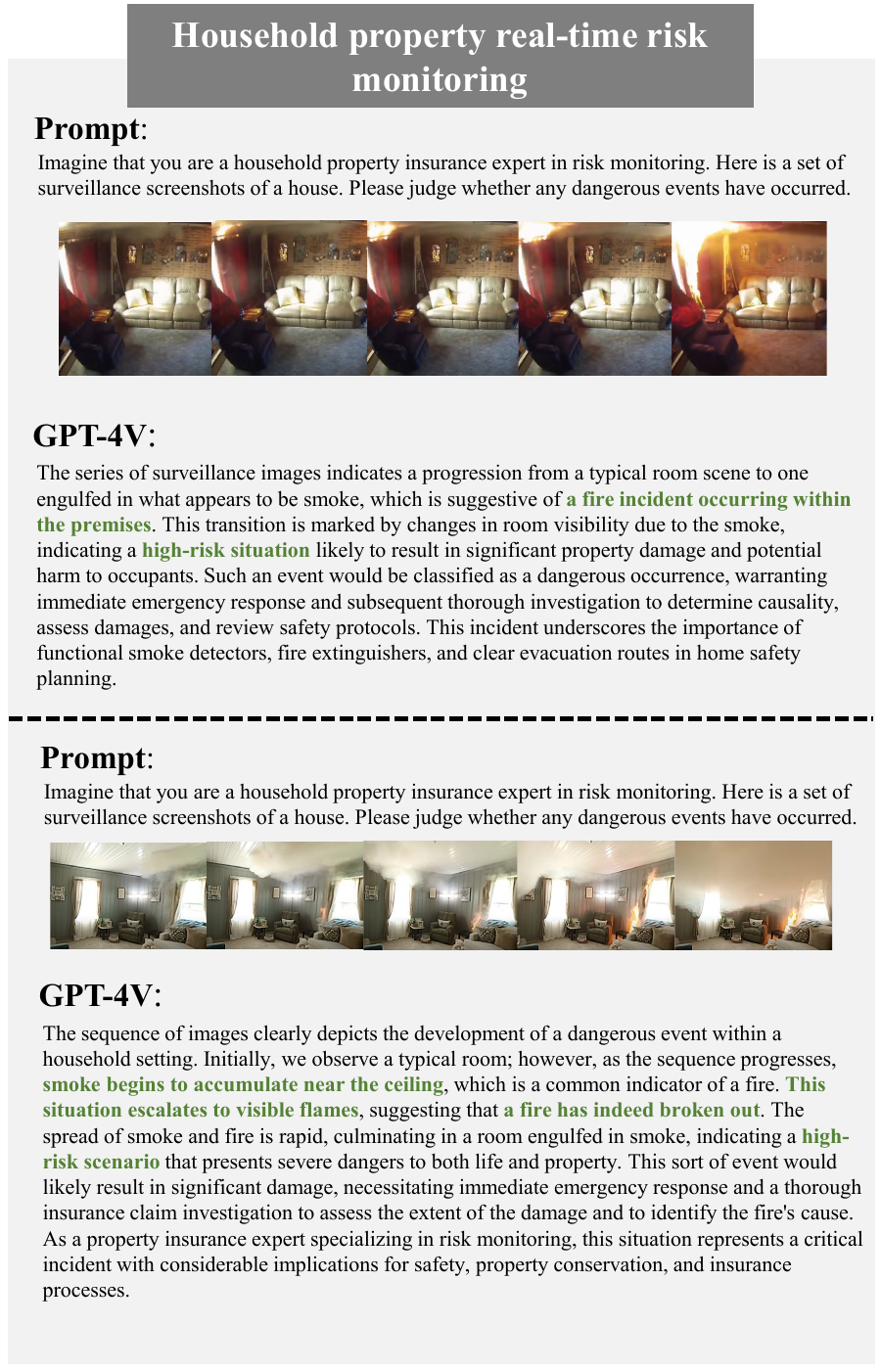} 
\caption[Section~\ref{sec: Household/Commercial Property Real-time Risk Monitoring}: household property real-time risk monitoring test case 1]{The test case of GPT-4V's capability in household property real-time risk monitoring. The relevant parts are highlighted in \textcolor[RGB]{84,130,53}{\textbf{green}}.}
\label{Household real-time risk monitoring} 
\end{figure}

We design two tasks to examine the risk monitoring capability of GPT-4V. The first task focuses on risk monitoring of dangerous events within a household. The prompt is: “\textit{Imagine that you are a household property insurance expert in risk monitoring. Here is a set of surveillance screenshots of a house. Please judge whether any dangerous events have occurred.}” The results (see Figure~\ref{Household real-time risk monitoring}) suggest that GPT-4V can accurately identify and detect unusual events (\eg fire) within a home and judge the trend of the fire situation (\eg “\textit{... smoke begins to accumulate ...}”, “\textit{... escalates to visible flames ...}”).

\begin{figure}
\centering 
\includegraphics[width=1\textwidth,height=1.45\textwidth]{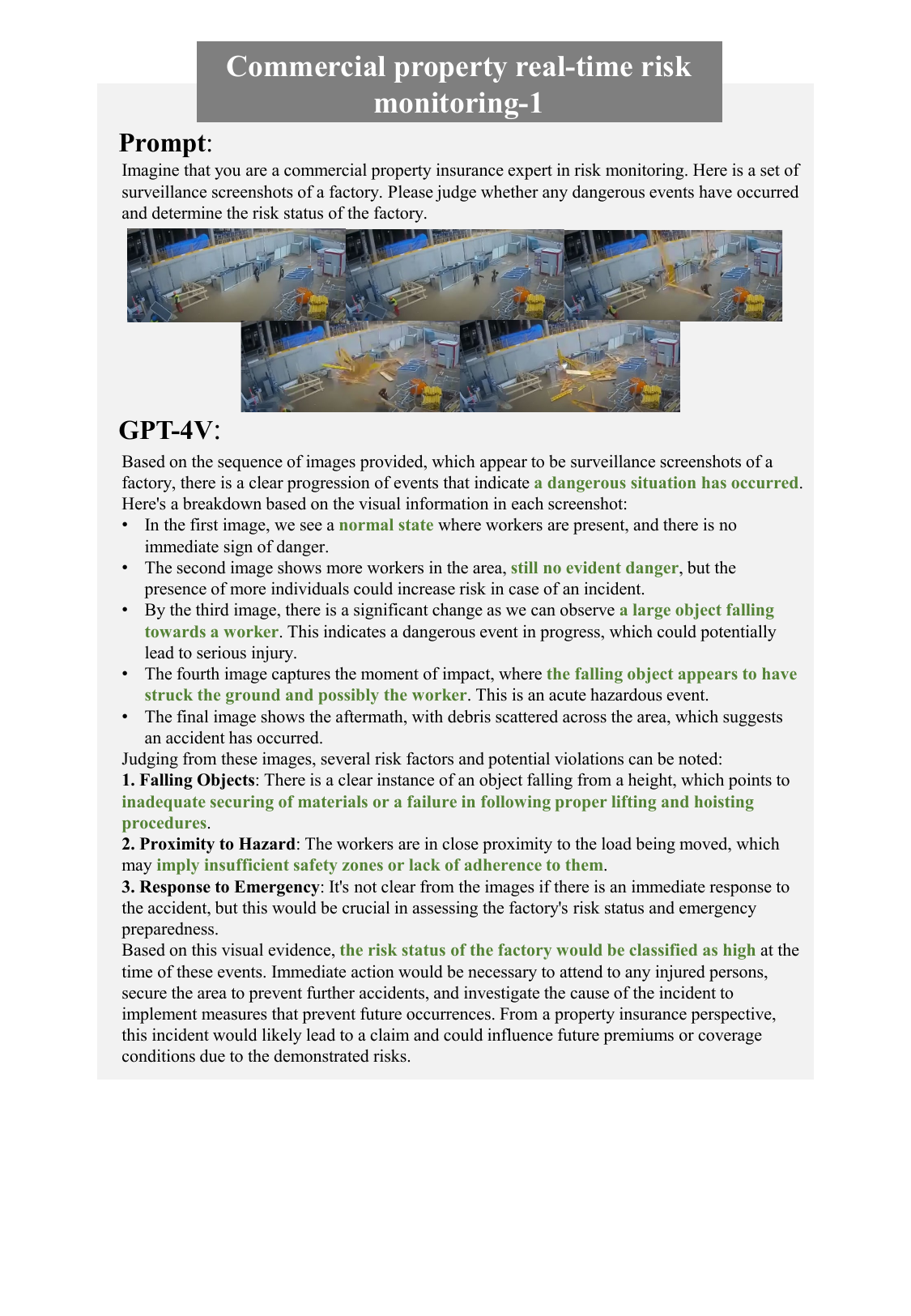} 
\caption[Section~\ref{sec: Household/Commercial Property Real-time Risk Monitoring}: commercial property real-time risk monitoring test case 1]{The test case of GPT-4V's capability in commercial property real-time risk monitoring. The relevant parts are highlighted in \textcolor[RGB]{84,130,53}{\textbf{green}}.}
\label{Commercial property real-time risk monitoring-1} 
\end{figure}

\begin{figure}
\centering 
\includegraphics[width=1\textwidth,height=1.45\textwidth]{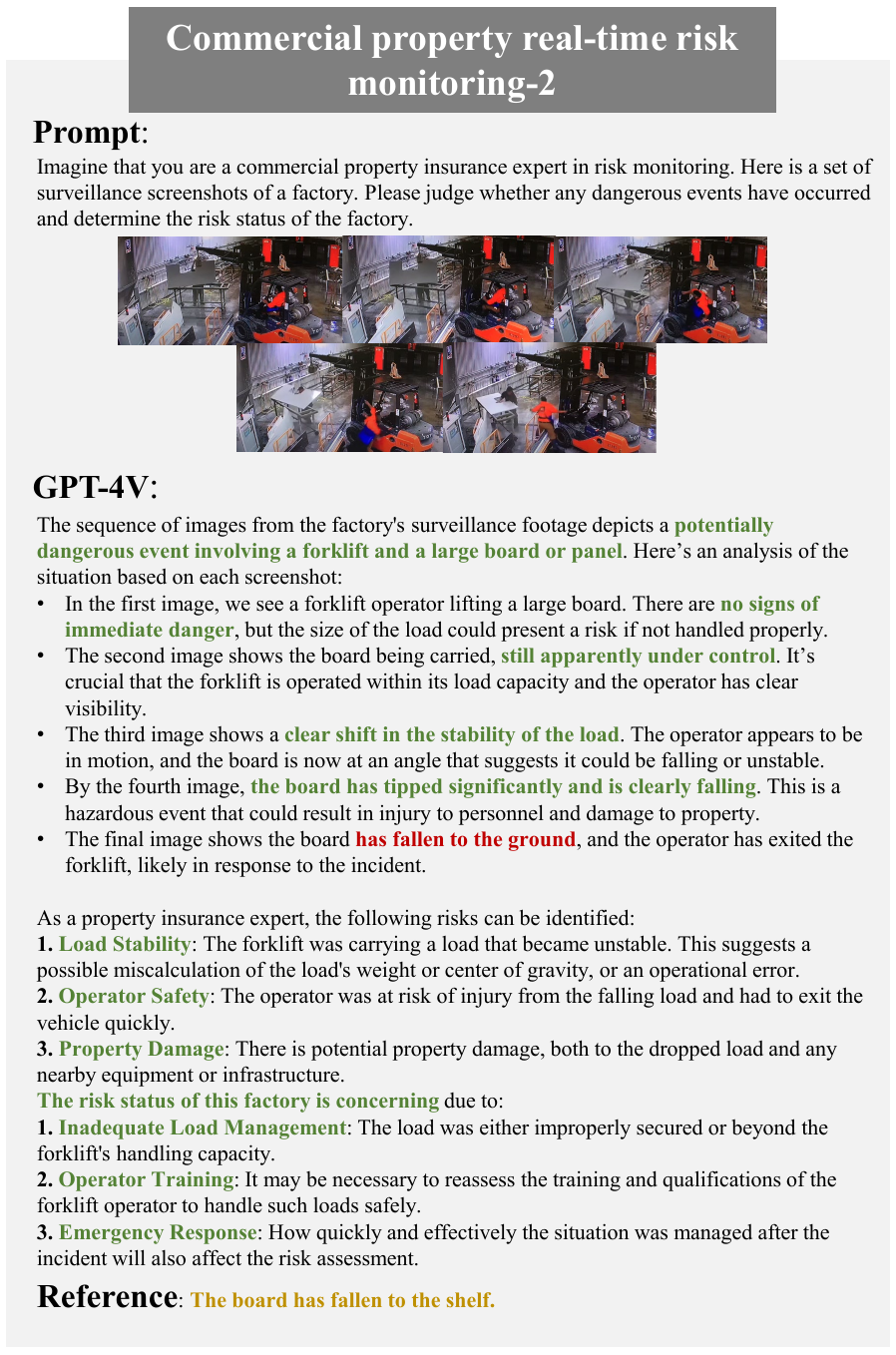} 
\caption[Section~\ref{sec: Household/Commercial Property Real-time Risk Monitoring}: commercial property real-time risk monitoring test case 2]{The test case of GPT-4V's capability in commercial property real-time risk monitoring. The correct answer parts, incorrect answer parts, and references are highlighted in \textcolor[RGB]{84,130,53}{\textbf{green}}, \textcolor[RGB]{192,0,0}{\textbf{red}} and \textcolor[RGB]{191,144,0}{\textbf{yellow}}, respectively.}
\label{Commercial property real-time risk monitoring-2} 
\end{figure}

The second task involves risk monitoring of hazardous events within a factory and analyzing the risk status. The prompt is: “\textit{Imagine that you are a commercial property insurance expert in risk monitoring. Here is a set of surveillance screenshots of a factory. Please judge whether any dangerous events have occurred and determine the risk status of the factory.}” The experimental results (see Figures~\ref{Commercial property real-time risk monitoring-1} and \ref{Commercial property real-time risk monitoring-2}) demonstrate that GPT-4V can accurately identify dangerous events in a factory (\eg operational errors, high-altitude falls) and assess the risk status (\eg “\textit{... The risk status of this factory is concerning ...}”).

\subsubsection{Household and Commercial Property Claims Processing}\label{sec: Household/Commercial property damage evaluation}

Claims processing in household and commercial property insurance involves critically evaluating the damage to buildings and property to determine the cause, extent, and associated cost~\cite{saeki2000seismic,dong2013comprehensive,rajagopal2020review}. Computer vision can significantly aid this task—for instance, by using drones to capture photos of a house's roof and employing computer vision to assess the damage to the roof~\cite{hezaveh2017roof}, or by analysing post-disaster images of a house to determine the extent of damage sustained~\cite{nia2017building}. These technologies have greatly improved the efficiency of claims processing.

To assess GPT-4V's capabilities in claims processing, we design two tasks. The first task focuses on evaluating loss to household property, encompassing the analysis of the cause of loss, assessment of the damage level, and estimation of the loss. We use two different scenarios for this task: roof and interior damage. The prompt is: “\textit{Imagine that you are a household insurance expert in claims processing. Here is an image of  a house. Please briefly evaluate the damage, analyze the cause of damage and give a loss level and estimated loss.}”

The second task involves evaluating loss to commercial property, again including analysis of the cause of loss, assessment of the damage level, and estimation of the loss. The prompt is: “\textit{Imagine that you are a commercial property insurance expert in evaluating damage. Here is an image of damaged commercial facilities. Please briefly evaluate the damage, analyze the cause of damage and give a loss level and estimated loss.}”

\begin{figure}
\centering 
\includegraphics[width=1\textwidth,height=1.45\textwidth]{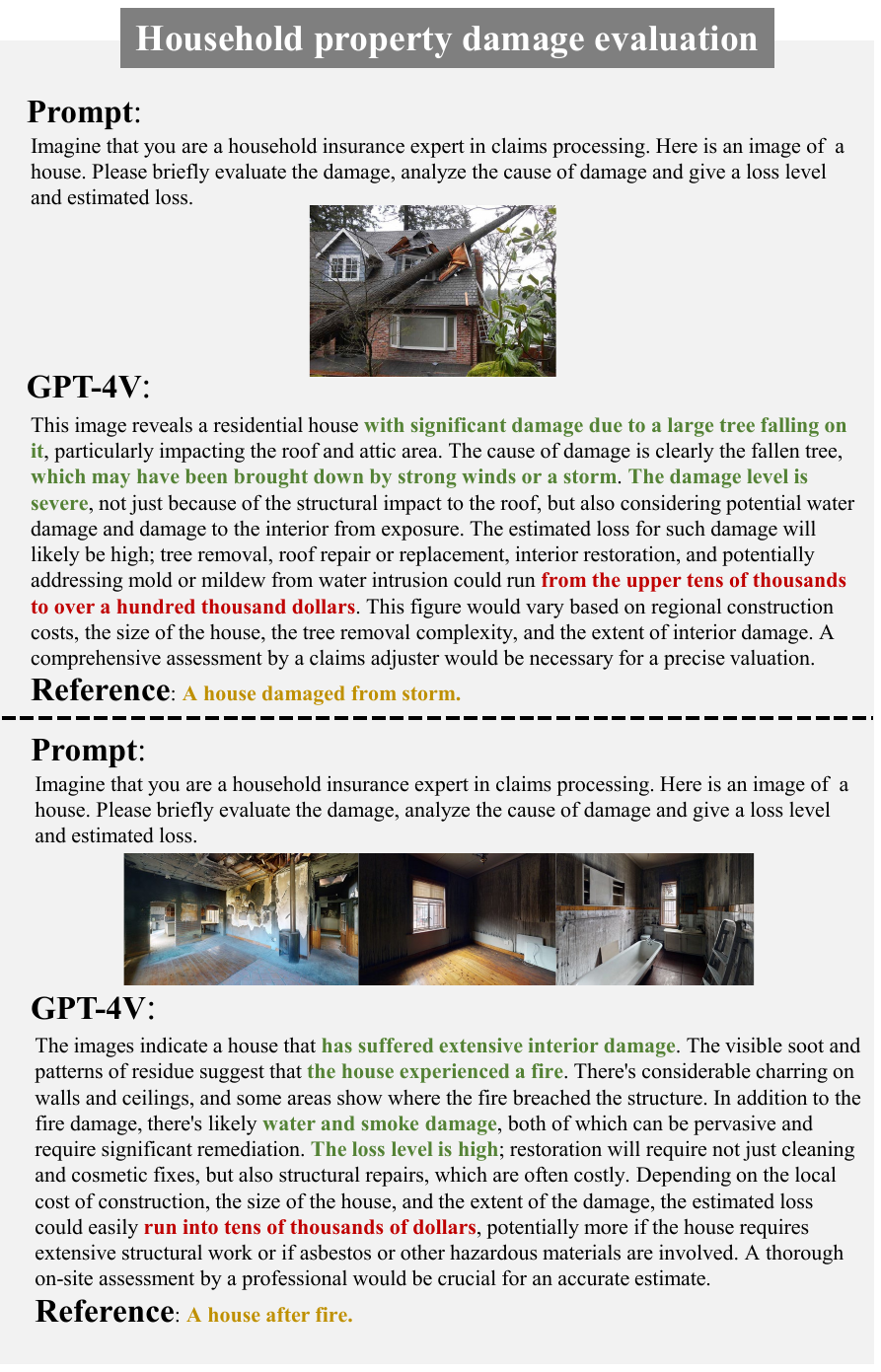} 
\caption[Section~\ref{sec: Household/Commercial property damage evaluation}: household property damage evaluation test case]{The test case of GPT-4V's capability in household property damage evaluation. The correct answer parts, incorrect answer parts, and references are highlighted in \textcolor[RGB]{84,130,53}{\textbf{green}}, \textcolor[RGB]{192,0,0}{\textbf{red}} and \textcolor[RGB]{191,144,0}{\textbf{yellow}}, respectively.}
\label{household property damage evaluation} 
\end{figure}

\begin{figure}
\centering 
\includegraphics[width=1\textwidth,height=1.45\textwidth]{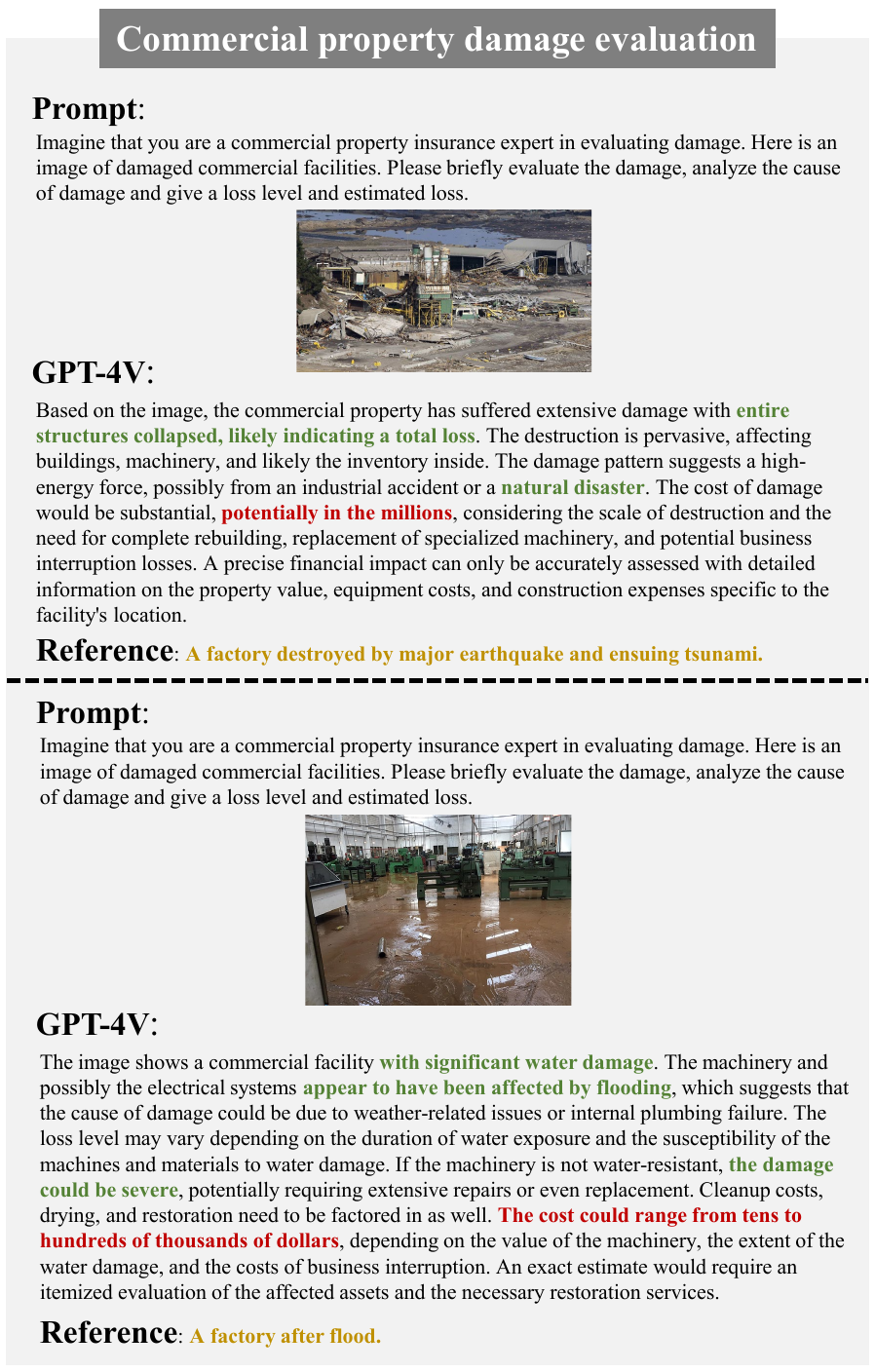} 
\caption[Section~\ref{sec: Household/Commercial property damage evaluation}: commercial property damage evaluation test case]{The test case of GPT-4V's capability in commercial property damage evaluation. The correct answer parts, incorrect answer parts, and references are highlighted in \textcolor[RGB]{84,130,53}{\textbf{green}}, \textcolor[RGB]{192,0,0}{\textbf{red}} and \textcolor[RGB]{191,144,0}{\textbf{yellow}}, respectively.}
\label{commercial property damage evaluation} 
\end{figure}

In both tasks, GPT-4V demonstrates its remarkable ability to evaluate damage (see Figures~\ref{household property damage evaluation} and \ref{commercial property damage evaluation}). It effectively outlines the extent of property damage (\eg “\textit{... with significant damage due to a large tree falling on it ...}”), identifies the causes of damage (\eg storm, fire, flood), and assesses the severity of the damage (\eg “\textit{... The damage level is severe ...}”, “\textit{... The loss level is high ...}”). However, GPT-4V falls short in estimating the precise monetary value of the damage, providing only a rough range (\eg “\textit{... from the upper tens of thousands to over a hundred thousand dollars ...}”).

\begin{figure}
\centering 
\includegraphics[width=1\textwidth,height=1.56\textwidth]{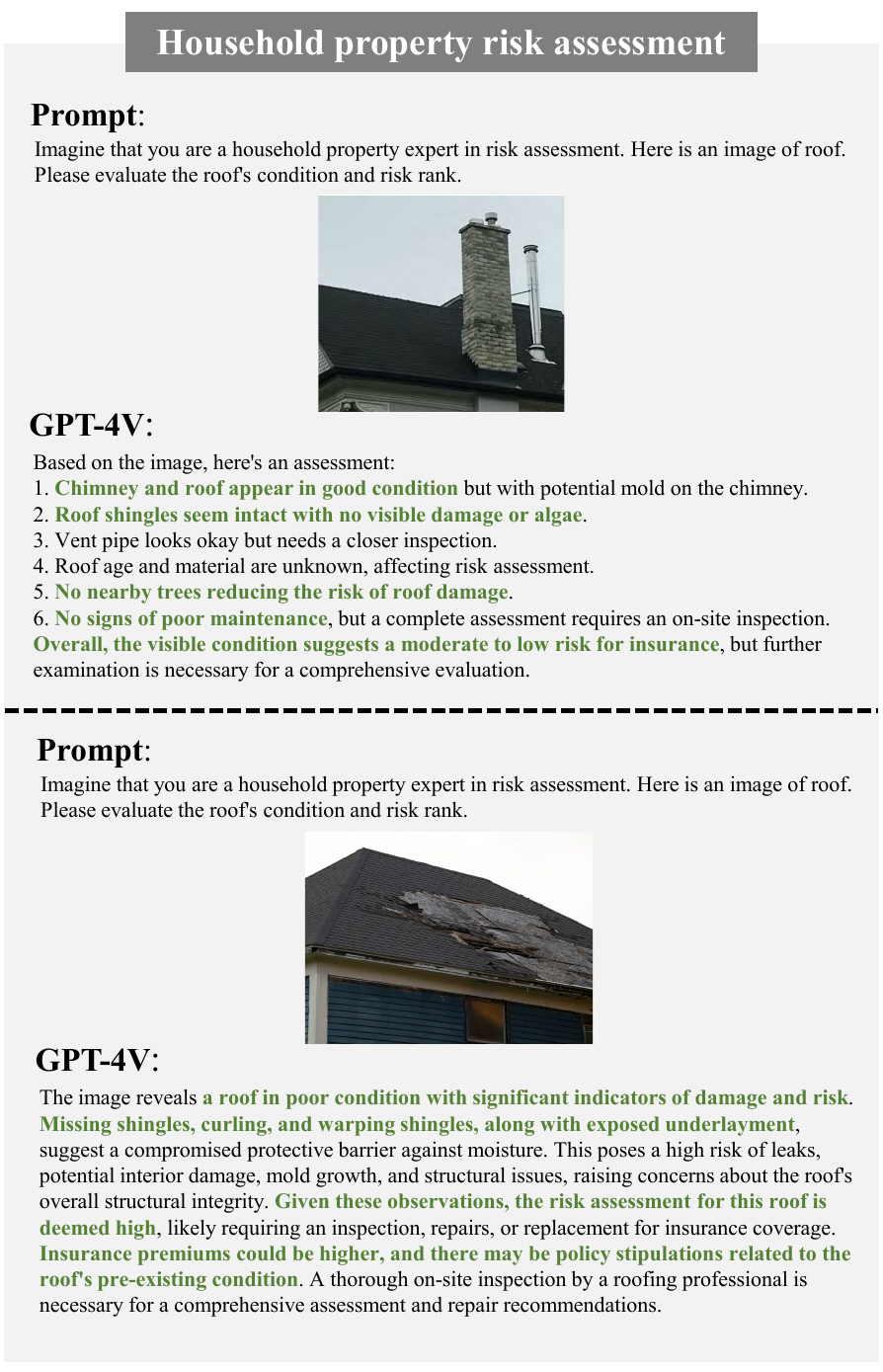} 
\caption[Section~\ref{sec: Household/Commercial Property Risk Assessment}: household property risk assessment test case]{The test case of GPT-4V's capability in household property risk assessment. The relevant parts are highlighted in \textbf{\textcolor[RGB]{84,130,53}{\textbf{green}}}.} 
\label{household property risk assessment} 
\end{figure}

\newpage
\subsection{Health Insurance}\label{sec:health}

Health insurance provides financial protection against health-related risks such as medical expenses, illness, and disability~\cite{levy2008impact}. The advent of computer vision technologies is transforming the health industry, enhancing medical procedures, accelerating healthcare research, and improving patient experiences~\cite{jiang2017artificial,gao2018computer,shaheen2021applications,zhang2021monitoring,lyu2022social,fan2021american}. This transformation significantly impacts the health insurance sector by enhancing work efficiency, improving service quality, and reducing errors and fraud~\cite{doss2022intelligent,doss2042health}. While some research has explored GPT-4V's capabilities in the medical field~\cite{wu2023can,ferber2024context,nori2023capabilities}, focusing mainly on assisting with the diagnosis of medical images (such as X-rays, MRIs, \textit{etc}.), our study concentrates on image tasks related to health insurance. We cover three main areas: health risk assessment, health risk monitoring, and assistance with health claims.

\subsubsection{Health Risk Assessment}\label{sec: Health Risk Assessment}

Before underwriting a health insurance policy, insurers must evaluate the insured's health condition. This entails a thorough review of the insured's medical history, examination reports, and other relevant documents to accurately assess health risks, a process that can be time-consuming~\cite{gleeson2009medical,pnevmatikakis2021risk}. The adoption of computer vision could significantly enhance the precision and efficiency of this procedure~\cite{tafti2016ocr,kumar2022ocr}.

To test GPT-4V's capabilities in health risk assessment, we design a task involving medical document extraction and health risk analysis. We select three distinct types of medical documents (a histopathology report, a discharge report, and a physical examination report) and provide the following prompt: “\textit{Imagine that you are a health insurance expert in risk assessment. Here is an image of a patient's medical report. Please extract the text, briefly evaluate the health risk, and assign a risk level.}” The experimental results (see Figures~\ref{health risk assessment-1}, \ref{health risk assessment-2}, and \ref{health risk assessment-3}) indicate that GPT-4V's text extraction capabilities are influenced by the language of the document. It demonstrates high precision in extracting English text but struggles with Chinese text, aligning with the findings of Shi~\etal~\cite{shi2023exploring}. For the case presented in Figure ~\ref{health risk assessment-3}, we translate the examination report into English before inputting it into GPT-4V. In terms of health risk assessment based on medical document content, GPT-4V shows strong analytical and evaluative abilities, accurately assessing the insured's risk status (\eg “\textit{... the patient would be considered at a high health risk ...}”, “\textit{... overweight status increases the client's risk for numerous chronic conditions ...}”).

\subsubsection{Health Risk Monitoring}\label{sec: Health Risk Monitoring}

In recent years, remote monitoring technology has been increasingly deployed in fields such as healthcare and insurance, enabling real-time monitoring of patients' health risk statuses through contact-based devices or non-contact video surveillance~\cite{malasinghe2019remote}. For example, insurance companies can monitor individuals' routine physiological metrics using contact-based instruments like blood pressure monitors and pulse oximeters, and they can also detect and issue timely alerts for abnormal events, such as falls, using home surveillance cameras~\cite{han2022proposal}. With the aid of computer vision technology, these monitoring methods can achieve automation and intelligence~\cite{rougier2011robust,boonnag2023pacman}.

\begin{figure}
\centering 
\includegraphics[width=1\textwidth,height=1.45\textwidth]{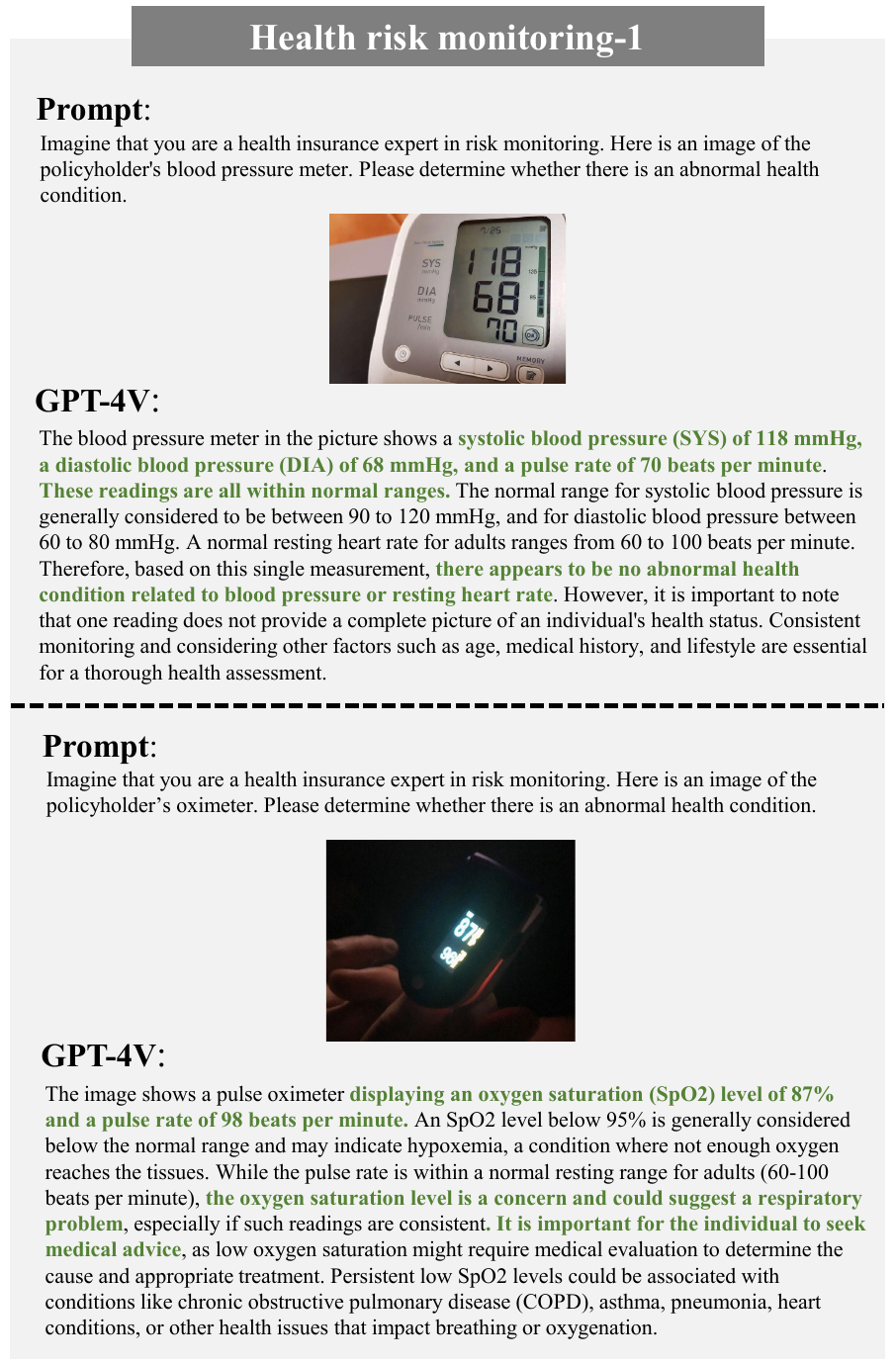} 
\caption[Section~\ref{sec: Health Risk Monitoring}: health risk monitoring test case 1]{The test case of GPT-4V's capability in health risk monitoring. The relevant parts are highlighted in \textcolor[RGB]{84,130,53}{\textbf{green}}.} 
\label{Health risk monitoring-1} 
\end{figure}

We design two tasks to test GPT-4V's capabilities in health risk monitoring: the first task involves information extraction and anomaly detection from health monitoring devices (blood pressure meter, oximeter). The prompt is: “\textit{Imagine that you are a health insurance expert in risk monitoring. Here is an image of the policyholder's blood pressure meter/oximeter. Please determine whether there is an abnormal health condition.}” The experimental results (see Figure~\ref{Health risk monitoring-1}) show that GPT-4V can effectively identify the readings of both types of health monitoring devices. Notably, in the case of the oximeter, where it is challenging for the human eye to discern which readings correspond to which metrics, GPT-4V accurately identifies them (“\textit{... displaying an oxygen saturation (SpO2) level of 87\% and a pulse rate of 98 beats per minute ...}”). Furthermore, GPT-4V is capable of identifying and detecting abnormal health conditions based on the device readings (\eg “\textit{... the oxygen saturation level is a concern and could suggest a respiratory problem ...}”).

\begin{figure}
\centering 
\includegraphics[width=1\textwidth,height=1.45\textwidth]{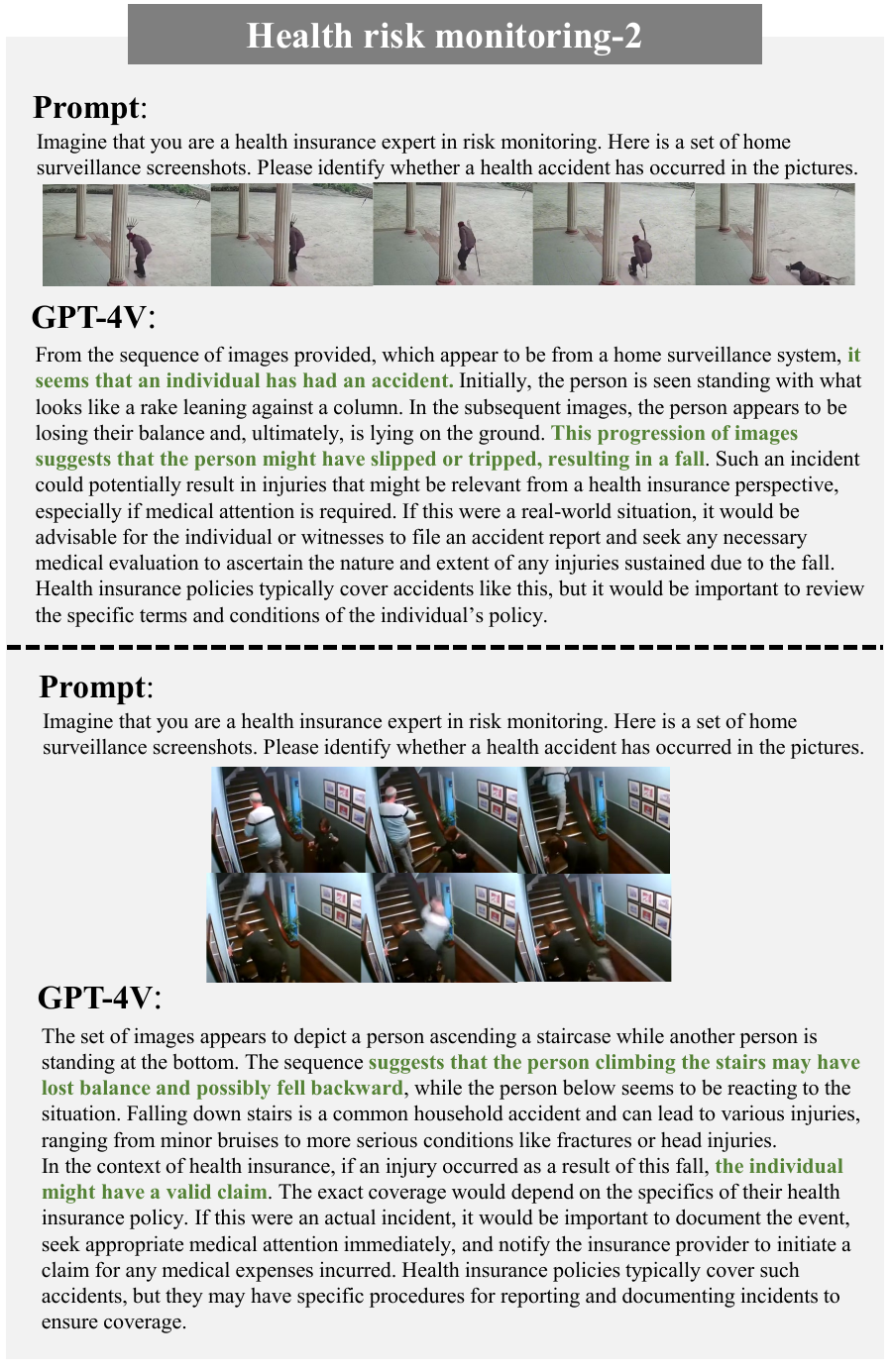} 
\caption[Section~\ref{sec: Health Risk Monitoring}: health risk monitoring test case 2]{The test case of GPT-4V's capability in health risk monitoring. The relevant parts are highlighted in \textcolor[RGB]{84,130,53}{\textbf{green}}.} 
\label{Health risk monitoring-2} 
\end{figure}

The second task involves real-time health risk detection through monitoring devices, using second-by-second frames from home surveillance as input. The prompt is: “\textit{Imagine that you are a health insurance expert in risk monitoring. Here is a set of home surveillance screenshots. Please identify whether a health accident has occurred in the pictures.}” The results (see Figure~\ref{Health risk monitoring-2}) demonstrate that GPT-4V can accurately identify events in the frames that pose health hazards (\eg “\textit{... suggests that the person might have slipped or tripped, resulting in a fall ...}”), and provide insurance recommendations based on the events depicted (\eg “\textit{... the individual might have a valid claim ...}”).

\subsubsection{Health Insurance Claims Processing}\label{sec: Health Insurance Claims Assistance}

Processing health insurance claims involves verifying medical consultation materials submitted by policyholders, such as medical records, diagnostic results, and expense summaries~\cite{popowich2005using}. This process requires the extraction of key information and analysis for potential fraud, abuse, and errors to ensure authenticity and compliance~\cite{lu2020embedding}. Traditionally, claims processing has relied on manual review, which can be time-consuming~\cite{anchan2011evaluation}. However, integrating AI technologies like computer vision can automate the process and detect potential fraudulent activities, thereby improving efficiency and accuracy~\cite{friskasari2019implementation,singh2021use}.

\begin{figure}
\centering 
\includegraphics[width=1\textwidth,height=1.42\textwidth]{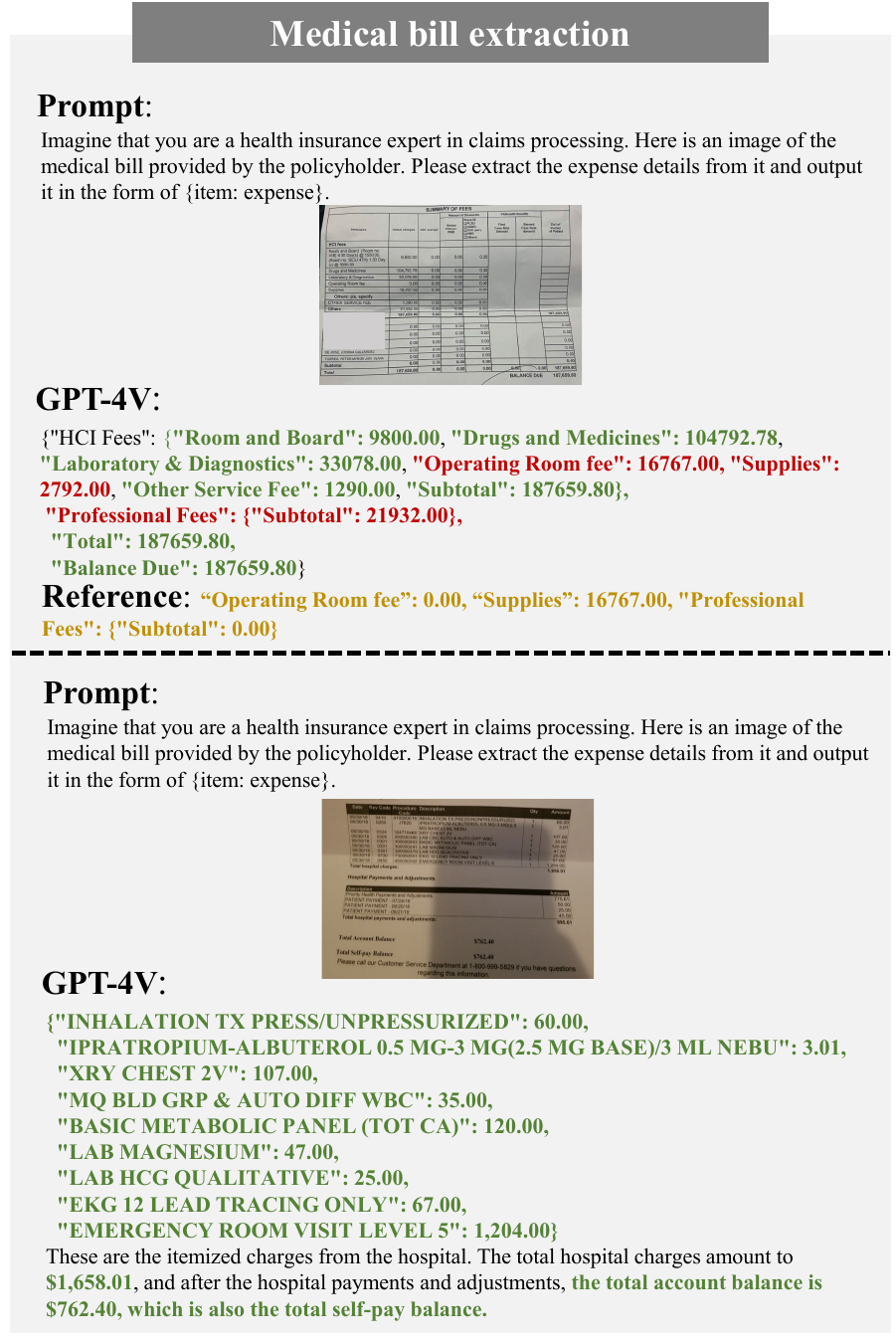} 
\caption[Section~\ref{sec: Health Insurance Claims Assistance}: medical bill extraction test case]{The test case of GPT-4V's capability in medical bill extraction. The correct answer parts, incorrect answer parts, and references are highlighted in \textcolor[RGB]{84,130,53}{\textbf{green}}, \textcolor[RGB]{192,0,0}{\textbf{red}} and \textcolor[RGB]{191,144,0}{\textbf{yellow}}, respectively.}
\label{Medical bill extraction} 
\end{figure}

We design two tasks to assess GPT-4V's capabilities in health insurance claims processing. The first task involves extracting expense items and details from the policyholder's medical expense bill, crucial for generating subsequent claims reports. The prompt is: “\textit{Imagine that you are a health insurance expert in claims processing. Here is an image of the medical bill provided by the policyholder. Please extract the expense details from it and output it in the form of \{item: expense\}.}” The experimental results (see Figure~\ref{Medical bill extraction}) indicate that while there are minor errors in GPT-4V's expense extraction (\eg Operating Room fee: 16767.00), the overall accuracy was relatively high, particularly for some key data (\eg Balance Due: 187659.80).

\begin{figure}
\centering 
\includegraphics[width=1\textwidth,height=1.42\textwidth]{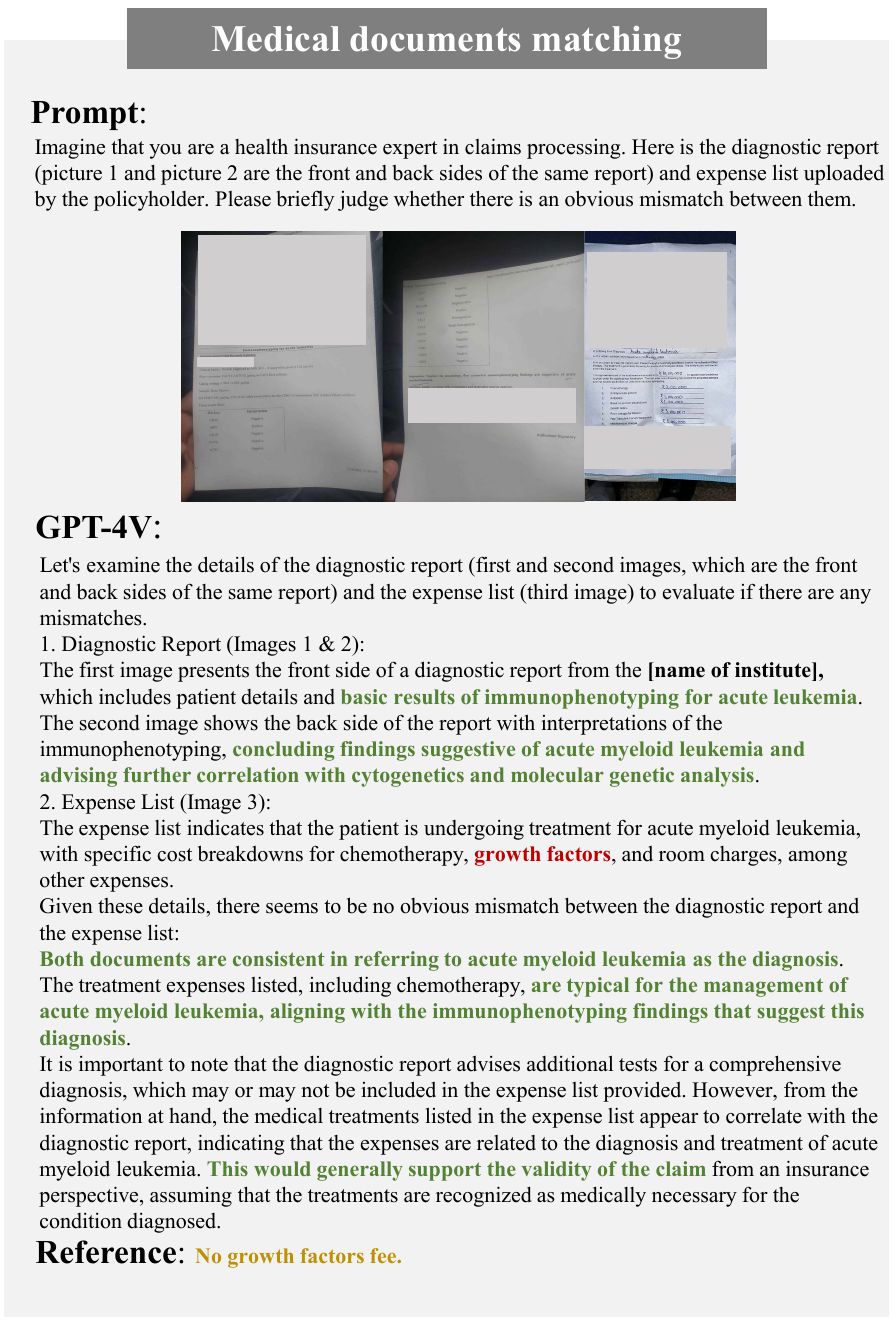} 
\caption[Section~\ref{sec: Health Insurance Claims Assistance}: medical documents matching test case]{The test case of GPT-4V's capability in medical documents matching. The correct answer parts, incorrect answer parts, and references are highlighted in \textcolor[RGB]{84,130,53}{\textbf{green}}, \textcolor[RGB]{192,0,0}{\textbf{red}} and \textcolor[RGB]{191,144,0}{\textbf{yellow}}, respectively.}
\label{Medical documents matching} 
\end{figure}

The second task involves a comparative analysis between the diagnostic report and the expense list to verify the relevance of medical items and the reasonableness of the claim. The prompt is: “\textit{Imagine that you are a health insurance expert in claims processing. Here is the diagnostic report and expense list uploaded by the policyholder. Please briefly judge whether there is an obvious mismatch between them.}” The experimental results (see Figure~\ref{Medical documents matching}) demonstrate that GPT-4V can accurately interpret the diagnostic report and expense list (\eg “\textit{... concluding findings suggestive of acute myeloid leukemia ...}”) and correctly analyze the reasonableness and relevance between the two documents (\eg “\textit{... Both documents are consistent in referring to acute myeloid leukemia as the diagnosis ...}”).

\begin{figure}
\centering 
\includegraphics[width=1\textwidth,height=1.3\textwidth]{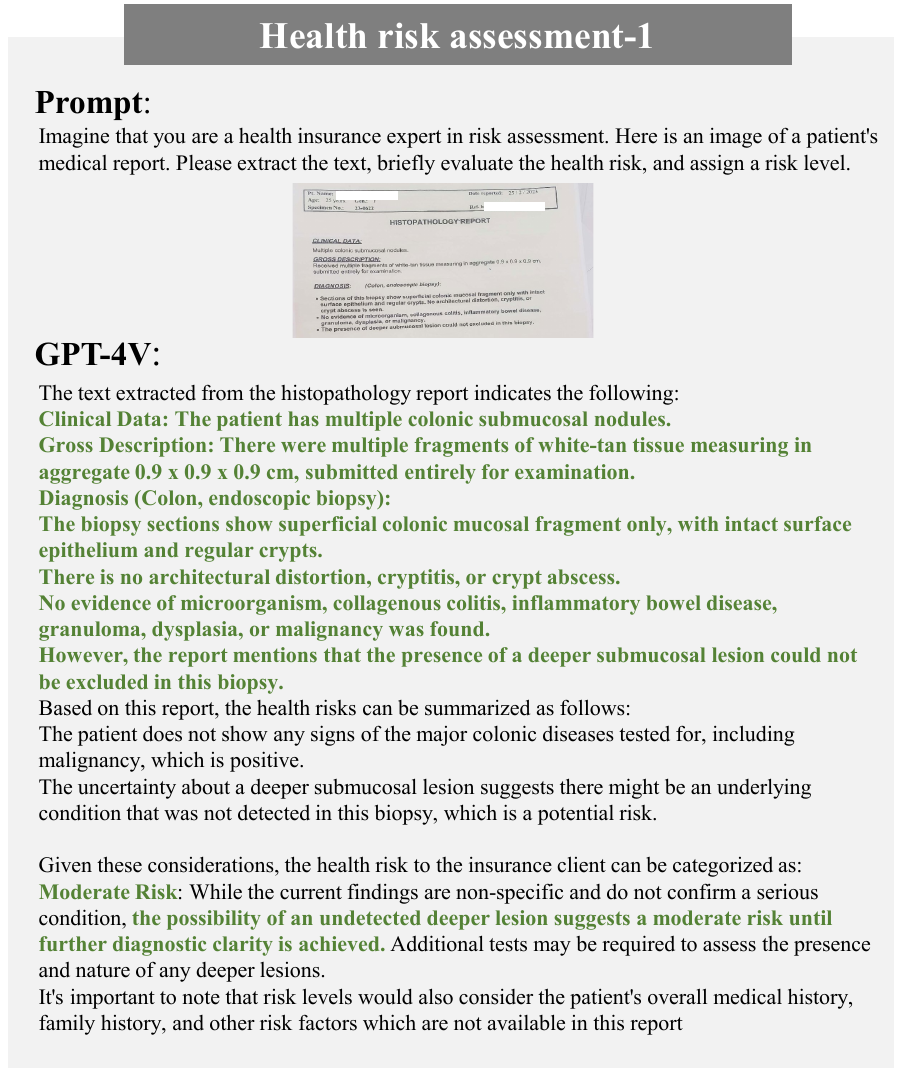} 
\caption[Section~\ref{sec: Health Risk Assessment}: health risk assessment test case 1]{The test case of GPT-4V's capability in health risk assessment. The relevant parts are highlighted in \textcolor[RGB]{84,130,53}{\textbf{green}}.} 
\label{health risk assessment-1} 
\end{figure}

\begin{figure}
\centering 
\includegraphics[width=1\textwidth,height=1.4\textwidth]{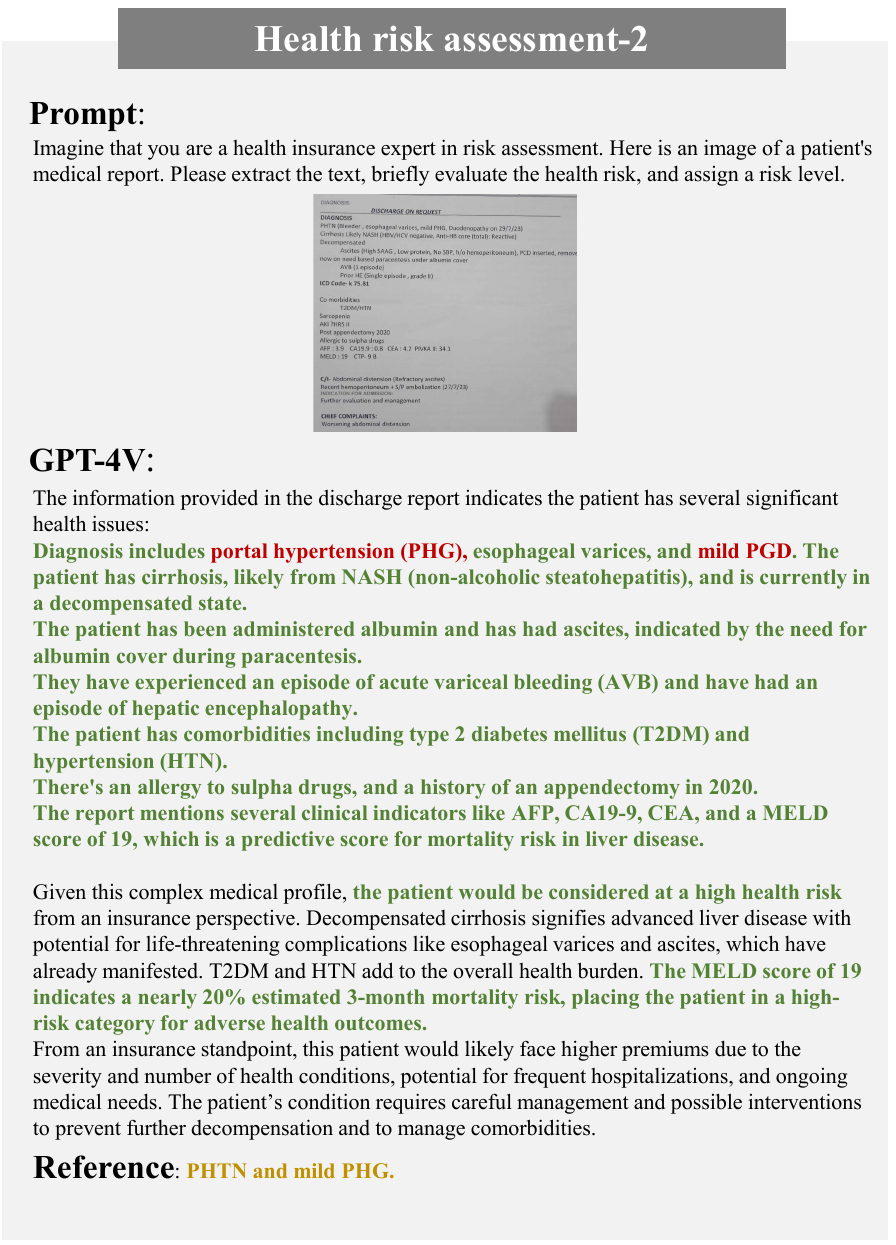} 
\caption[Section~\ref{sec: Health Risk Assessment}: health risk assessment test case 2]{The test case of GPT-4V's capability in health risk assessment. The correct answer parts, incorrect answer parts, and references are highlighted in \textcolor[RGB]{84,130,53}{\textbf{green}}, \textcolor[RGB]{192,0,0}{\textbf{red}} and \textcolor[RGB]{191,144,0}{\textbf{yellow}}, respectively.}
\label{health risk assessment-2} 
\end{figure}

\begin{figure}
\centering 
\includegraphics[width=1\textwidth,height=1.45\textwidth]{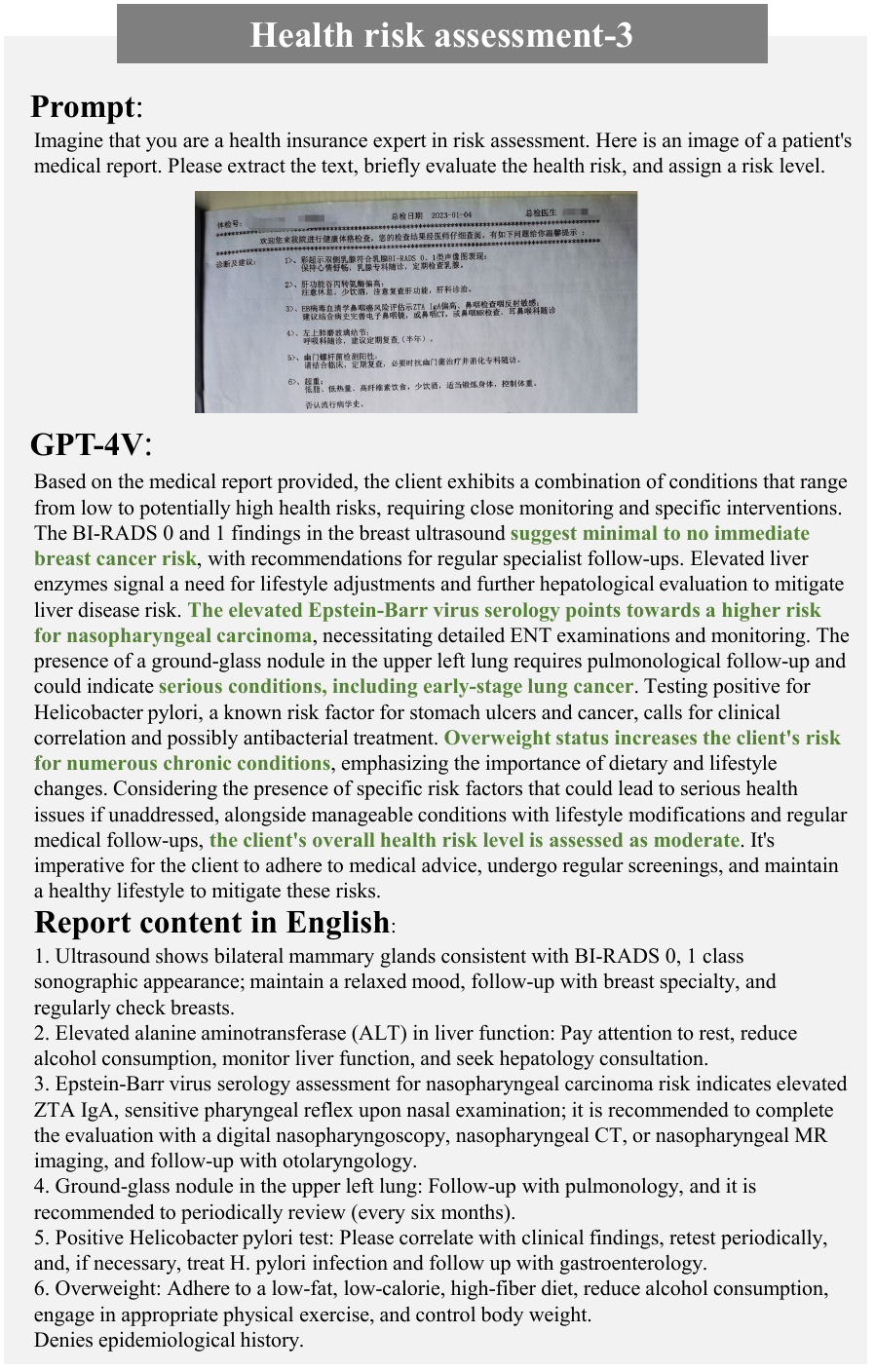} 
\caption[Section~\ref{sec: Health Risk Assessment}: health risk assessment test case 3]{The test case of GPT-4V's capability in health risk assessment. The relevant parts are highlighted in \textcolor[RGB]{84,130,53}{\textbf{green}}.} 
\label{health risk assessment-3} 
\end{figure}

\newpage

\subsection{Agricultural Insurance}\label{sec:agri}

Agriculture is a critical component of the primary sector and is intrinsically linked to people worldwide. Agricultural insurance plays an essential role in safeguarding farmers and agricultural operators from the impacts of natural disasters and market fluctuations~\cite{iturrioz2009agricultural}. It is pivotal in promoting agricultural development, stabilizing food prices, and ensuring food security. In recent years, the rapid advancement of smart agriculture has led to the emergence of computer vision detection systems as essential tools in agricultural operations~\cite{rico2019contextualized}. For instance, the use of drones and satellites to capture images of farmlands, combined with various artificial intelligence algorithms, provides functionalities such as risk monitoring, soil fertility testing, and crop growth surveillance~\cite{kurkute2018drones,brook2020smart}. These technologies can also enhance various aspects of agricultural insurance, such as yield prediction~\cite{singh2019tool} and fraud detection~\cite{sahni2020insurance}. While researchers~\cite{tan2023promises} have conducted preliminary explorations into the capabilities of GPT-4V in agriculture, their research primarily focuses on general agricultural scenarios. Our study, on the other hand, focuses on the application of agricultural insurance and tests the capabilities of GPT-4V across three dimensions: farmland risk assessment, farmland risk monitoring, and farmland claims processing.

\subsubsection{Farmland Risk Assessment}\label{sec: Farmland Risk Assessment}

Farmland risk assessment involves a comprehensive investigation of key risk factors such as crop types, farmland environment, water resource management, pest and disease presence, and natural climatic conditions to thoroughly evaluate the risk status of agricultural fields~\cite{gajic2019risk,lyubchich2019insurance}. Given the vast expanse of farmland, onsite human detection often cannot survey potential risks within the interior of the fields. Hence, the use of drones or satellite imagery is considered an effective approach, providing high-resolution representations of the entire farmland and assisting in the facilitation of risk assessment processes~\cite{atwood2005big,benami2021uniting}.

\begin{figure}
\centering 
\includegraphics[width=1\textwidth,height=1.42\textwidth]{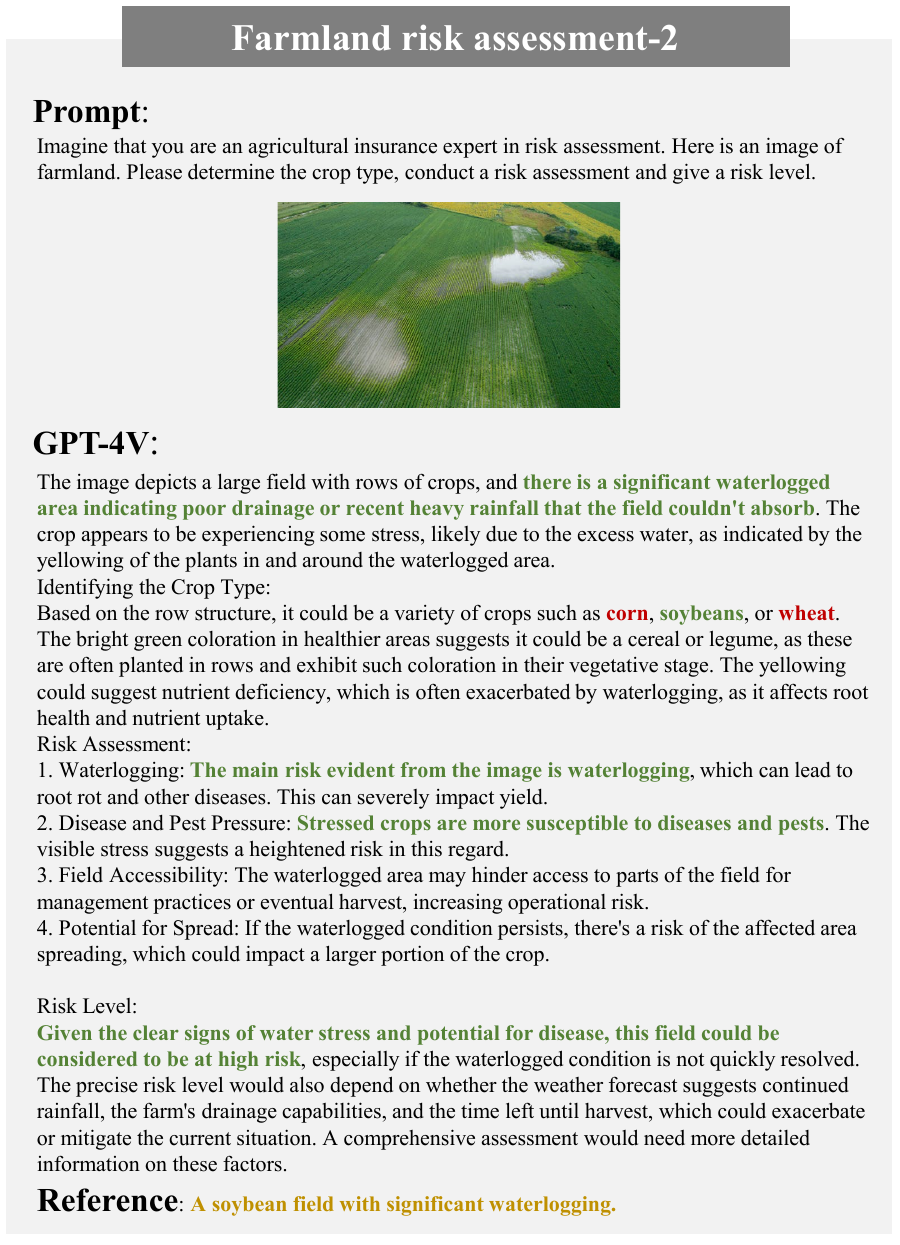} 
\caption[Section~\ref{sec: Farmland Risk Assessment}: farmland risk assessment test case 2]{The test case of GPT-4V's capability in farmland risk assessment. The correct answer parts, incorrect answer parts, and references are highlighted in \textcolor[RGB]{84,130,53}{\textbf{green}}, \textcolor[RGB]{192,0,0}{\textbf{red}} and \textcolor[RGB]{191,144,0}{\textbf{yellow}}, respectively.}
\label{Farmland risk assessment-2} 
\end{figure}

We design a task to assess GPT-4V's capabilities in farmland risk assessment. We select two drone-captured images of farmland and tasked GPT-4V with identifying crop varieties and evaluating associated risks. The prompt is: “\textit{Imagine that you are an agricultural insurance expert in risk assessment. Here is an image of farmland. Please determine the crop type, conduct a risk assessment and give a risk level.}” The test results (see Figures~\ref{Farmland risk assessment-1} and \ref{Farmland risk assessment-2}) reveal that GPT-4V is capable of identifying crop types from aerial imagery (although with some speculative elements in Figure~\ref{Farmland risk assessment-2}), providing accurate descriptions and analysis of the farmland's risk condition (\eg “\textit{... The health of the crops appears good ...}", “\textit{... The main risk evident from the image is waterlogging ...}”), and assessing the risk level of the farmland (\eg “\textit{... this field could be considered to be at high risk ...}”).

\subsubsection{Farmland Risk Monitoring}\label{sec: Farmland Risk Monitoring}

Farmland risk monitoring involves tracking crop growth conditions and identifying anomalous events that could impact agricultural output~\cite{lopez2018review,hafeez2022implementation}. Agriculture is a time-sensitive endeavor, with any unexpected incidents during crop growth phases potentially leading to significant losses, thus affecting both farmers and insurance companies financially. The use of drone and satellite imagery for real-time crop surveillance helps ensure stable crop development and mitigates risks~\cite{gulati2018crop,maimaitijiang2020crop}.

\begin{figure}
\centering 
\includegraphics[width=1\textwidth,height=1.5\textwidth]{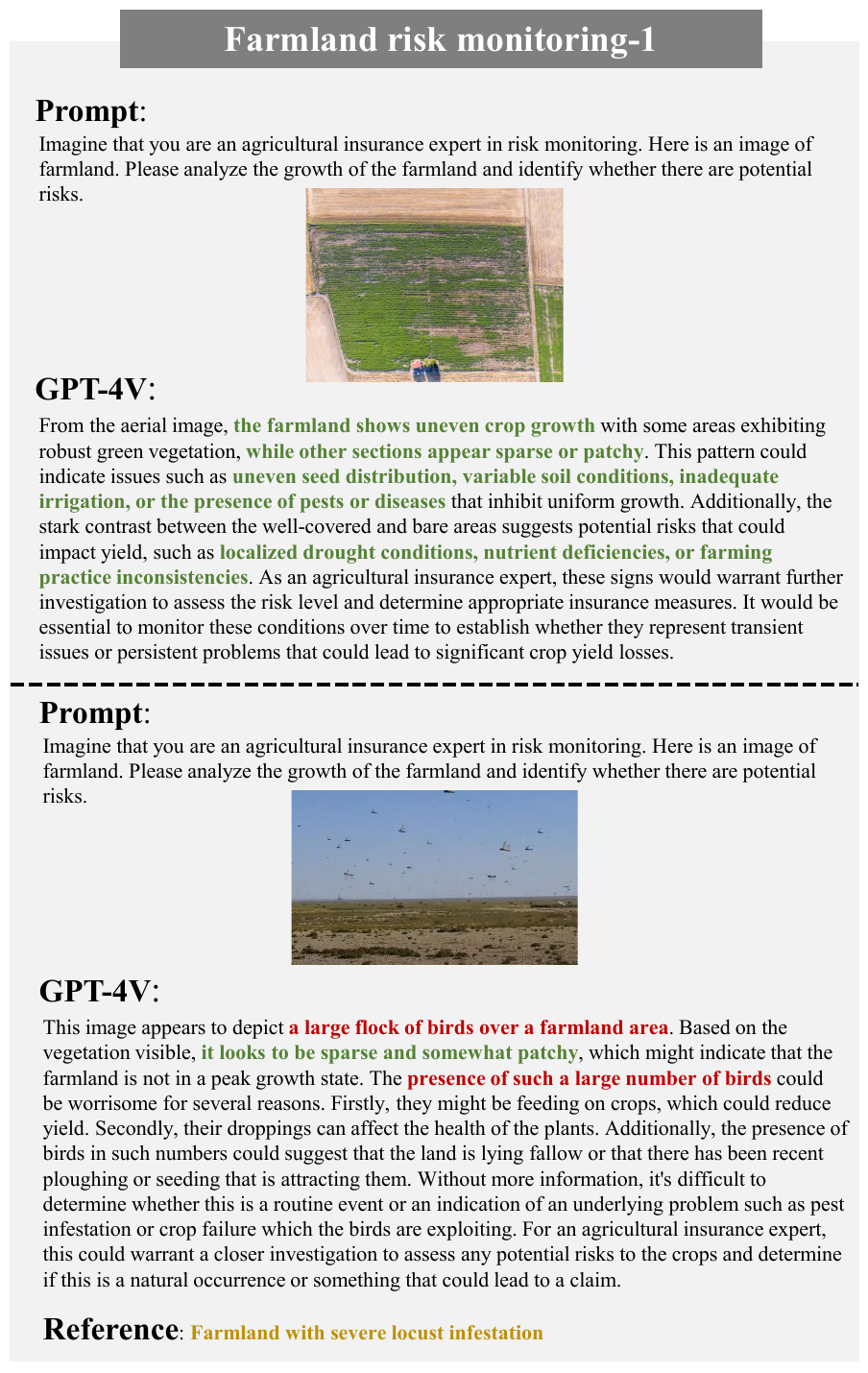} 
\caption[Section~\ref{sec: Farmland Risk Monitoring}: farmland risk monitoring test case 1]{The test case of GPT-4V's capability in farmland risk monitoring. The correct answer parts, incorrect answer parts, and references are highlighted in \textcolor[RGB]{84,130,53}{\textbf{green}}, \textcolor[RGB]{192,0,0}{\textbf{red}} and \textcolor[RGB]{191,144,0}{\textbf{yellow}}, respectively.}
\label{farmland risk monitoring-1} 
\end{figure}

\begin{figure}
\centering 
\includegraphics[width=1\textwidth,height=1.5\textwidth]{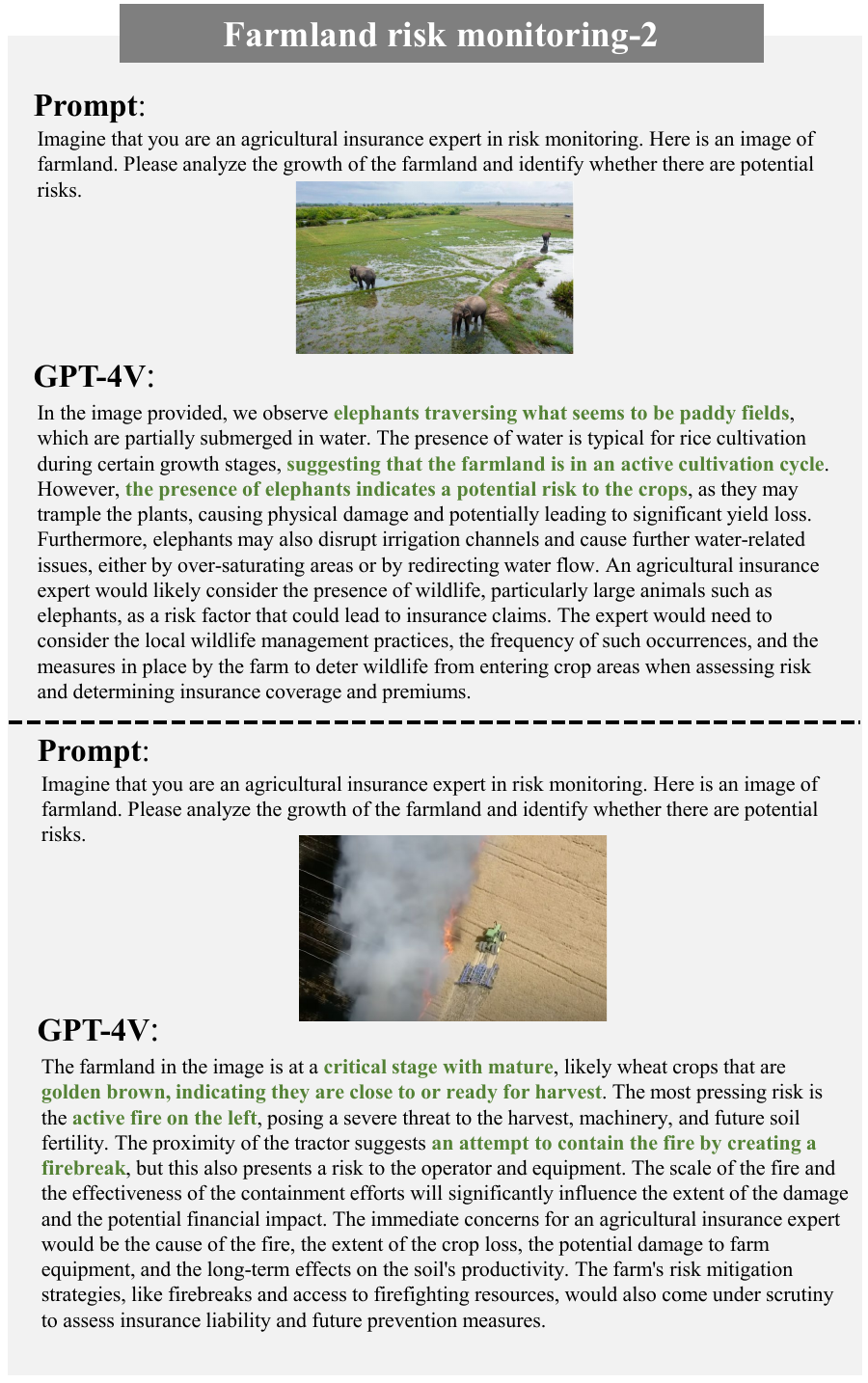} 
\caption[Section~\ref{sec: Farmland Risk Monitoring}: farmland risk monitoring test case 2]{The test case of GPT-4V's capability in farmland risk monitoring. The relevant parts are highlighted in \textcolor[RGB]{84,130,53}{\textbf{green}}.}
\label{farmland risk monitoring-2} 
\end{figure}

To evaluate GPT-4V's capabilities in farmland risk monitoring, we design a task that involves analyzing four scenarios featuring various anomalous events (\eg uneven crop growth, locust invasion, wildlife encroachment, agricultural fires). The task requires GPT-4V to examine the crop growth condition and detect any potential risks within the farmland. The prompt is: “\textit{Imagine that you are an agricultural insurance expert in risk monitoring. Here is an image of farmland. Please analyze the growth of the farmland and identify whether there are potential risks.}” The test results (see Figures~\ref{farmland risk monitoring-1} and \ref{farmland risk monitoring-2}) show that GPT-4V can provide accurate descriptions and analyses of crop growth (\eg “\textit{... the farmland shows uneven crop growth ...}”, “\textit{... indicating they are close to or ready for harvest ...}"). Moreover, it effectively identifies anomalous events (\eg “\textit{... the presence of elephants indicates a potential risk to the crops ...}”). However, there were minor recognition errors, such as in one case where a locust invasion was mistakenly identified as a large flock of birds (\eg “\textit{... a large flock of birds over a farmland area ...}”).

\subsubsection{Farmland Claims Processing}\label{sec: Farmland Damage Evaluation}

Farmland claims processing is a critical task that involves evaluating damage to determine the causes and assess the extent of the loss~\cite{chiu2020agriculture,iwahashi2022drought}. This process is essential for ensuring that farmers receive appropriate compensation and for streamlining the claims process. Insurance companies often use drones or smartphones to capture images of farmland, which are then combined with on-site inspections and image reviews for a comprehensive evaluation of the damage~\cite{gulati2018crop,ceballos2019feasibility}. Computer vision technologies can further enhance this process~\cite{tian2020computer}.

\begin{figure}
\centering 
\includegraphics[width=1\textwidth,height=1.5\textwidth]{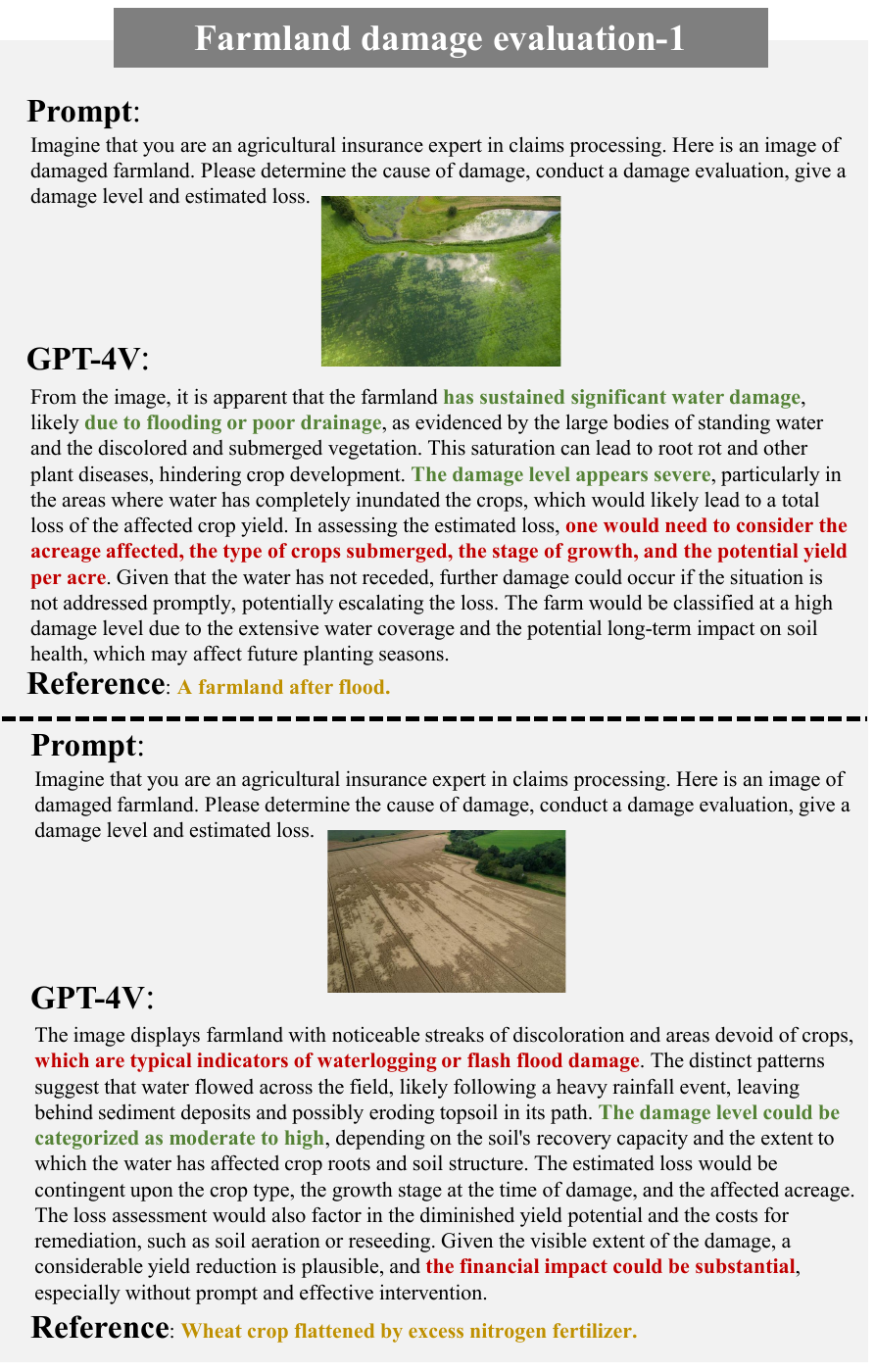} 
\caption[Section~\ref{sec: Farmland Damage Evaluation}: farmland damage evaluation test case 1]{The test case of GPT-4V's capability in farmland damage evaluation. The correct answer parts, incorrect answer parts, and references are highlighted in \textcolor[RGB]{84,130,53}{\textbf{green}}, \textcolor[RGB]{192,0,0}{\textbf{red}} and \textcolor[RGB]{191,144,0}{\textbf{yellow}}, respectively.}
\label{Farmland damage evaluation-1}
\end{figure}

\begin{figure}
\centering 
\includegraphics[width=1\textwidth,height=1.5\textwidth]{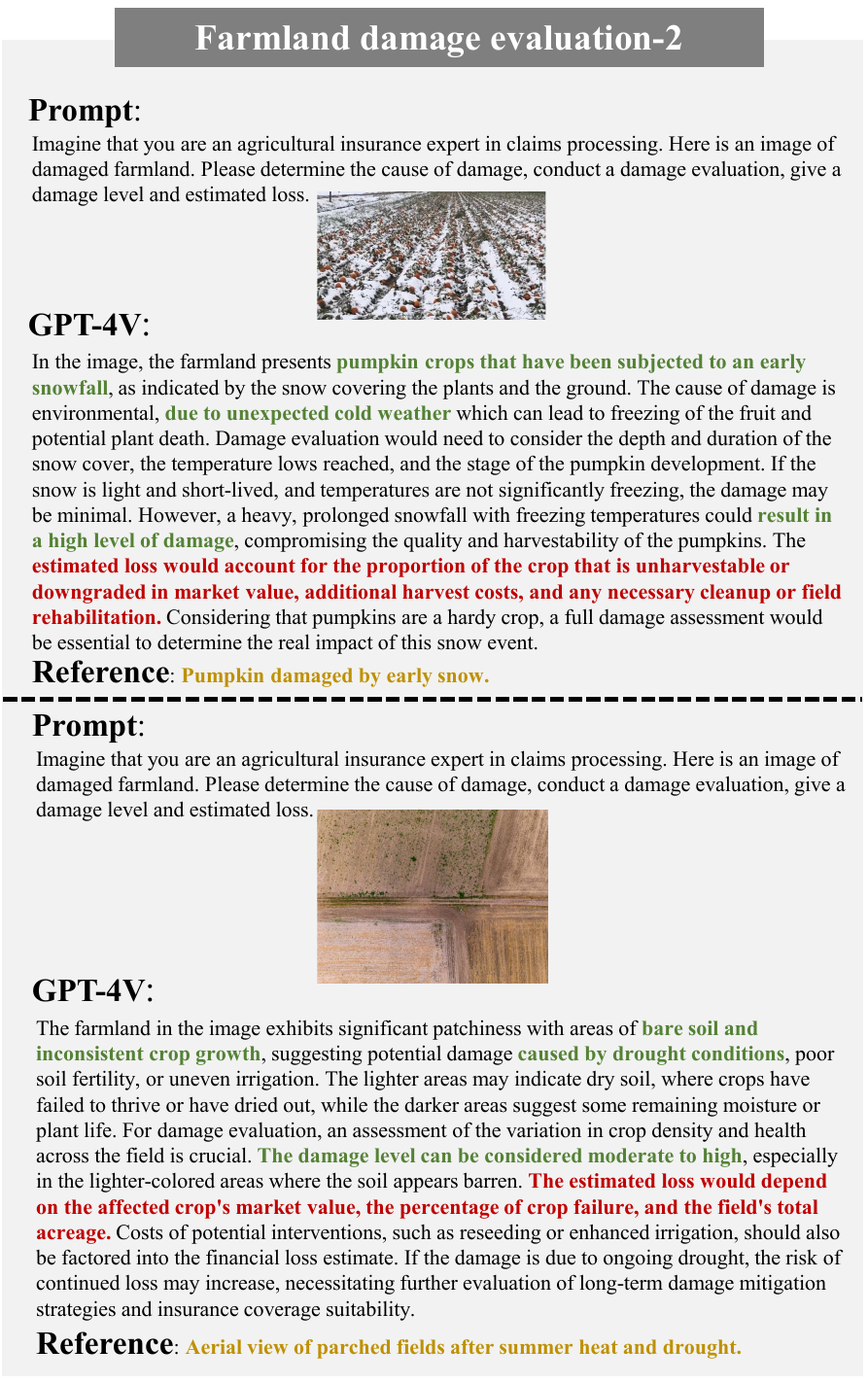} 
\caption[Section~\ref{sec: Farmland Damage Evaluation}: farmland damage evaluation test case 2]{The test case of GPT-4V's capability in farmland damage evaluation. The correct answer parts, incorrect answer parts, and references are highlighted in \textcolor[RGB]{84,130,53}{\textbf{green}}, \textcolor[RGB]{192,0,0}{\textbf{red}} and \textcolor[RGB]{191,144,0}{\textbf{yellow}}, respectively.}
\label{Farmland damage evaluation-2}
\end{figure}

To assess GPT-4V's capabilities in farmland damage evaluation, we design a task involving four different cases of farmland damage caused by various factors such as flooding, over-fertilization, early snow, and drought. We instruct GPT-4V to identify the causes of damage, assess the damage level, and estimate the scale of the loss. The prompt is: “\textit{Imagine that you are an agricultural insurance expert in claims processing. Here is an image of damaged farmland. Please determine the cause of damage, conduct a damage evaluation, give a damage level and estimated loss.}” The test results (see Figures~\ref{Farmland damage evaluation-1} and \ref{Farmland damage evaluation-2}) demonstrate that GPT-4V can accurately identify simpler and more evident causes of damage (\eg flooding, early snow, and drought), but it struggles with more complex scenarios (\eg over-fertilization). Additionally, it is capable of assessing the condition and level of damage (\eg “\textit{... The damage level could be categorized as moderate to high ...}”), yet it is unable to provide specific loss valuations (\eg “\textit{... The estimated loss would depend on the affected crop's market value, the percentage of crop failure ...}”).

\begin{figure}
\centering 
\includegraphics[width=1\textwidth,height=1.42\textwidth]{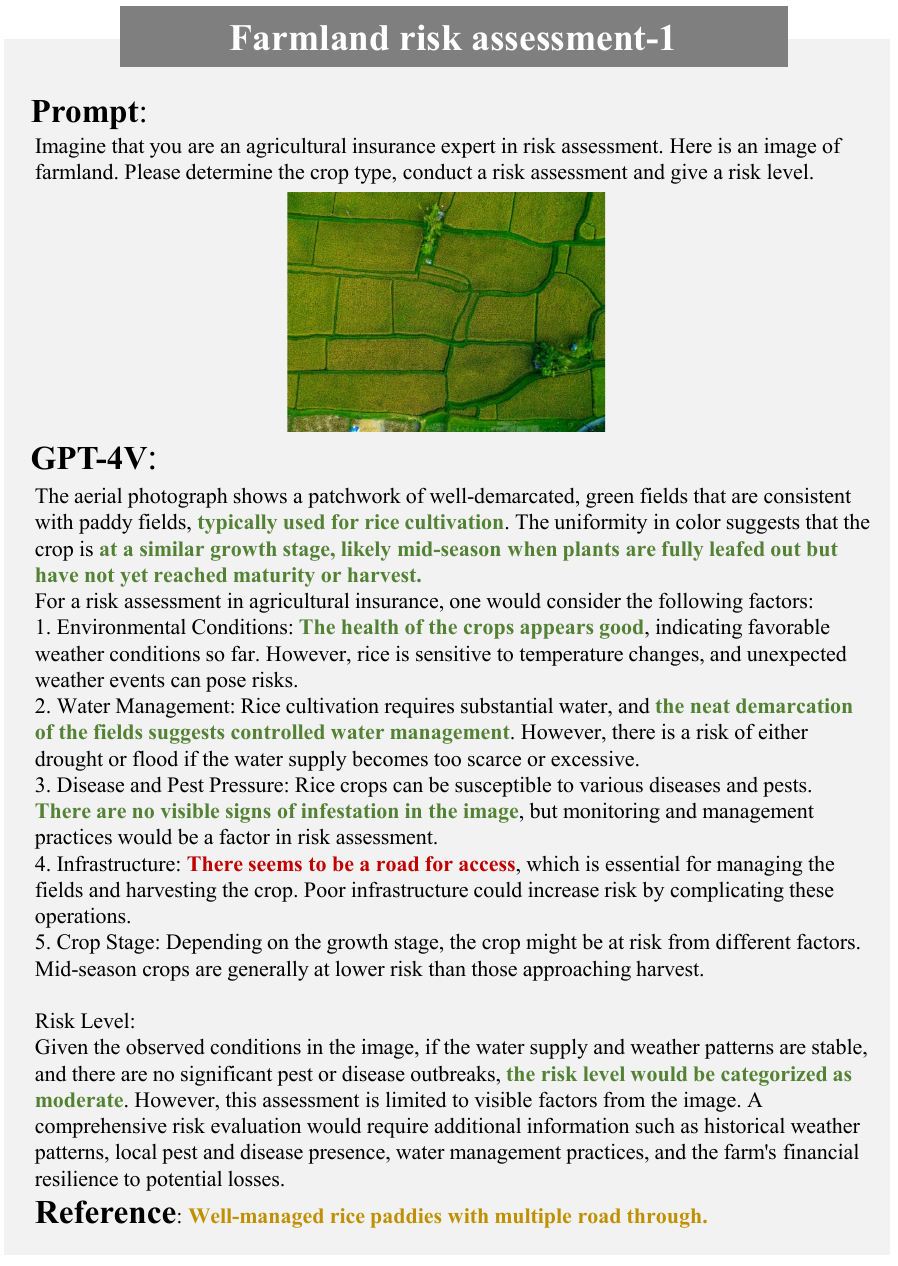} 
\caption[Section~\ref{sec: Farmland Risk Assessment}: farmland risk assessment test case 1]{The test case of GPT-4V's capability in farmland risk assessment. The correct answer parts, incorrect answer parts, and references are highlighted in \textcolor[RGB]{84,130,53}{\textbf{green}}, \textcolor[RGB]{192,0,0}{\textbf{red}} and \textcolor[RGB]{191,144,0}{\textbf{yellow}}, respectively.}
\label{Farmland risk assessment-1} 
\end{figure}

\newpage

\section{Challenges and Opportunities}\label{sec:Challenges and Opportunities}

\subsection{Accurate Damage Assessment}\label{sec: Accurate Damage Assessment}
We find that while GPT-4V is capable of conducting risk assessments and loss evaluations on a qualitative level, it encounters challenges with more detailed, quantitative rating and assessment tasks, such as making accurate predictions of risk probabilities and estimating the monetary extent of losses (\eg cases in Figures~\ref{extraction of auto accident elements test case}, \ref{household property risk assessment}, \ref{commercial property damage evaluation}, \ref{Farmland risk assessment-1} and \ref{Farmland damage evaluation-1}). This limitation does not fully meet the requirements of insurance professionals who often rely on precise numerical data for reference.  
To explore GPT-4V's capabilities in accurate damage assessment, we choose the car damage test case in Figure~\ref{extraction of auto accident elements test case} as an example, and use a new prompt approach for further exploration.

We construct a table listing the repair costs for parts of two car models, Honda Fit and Audi A6, with cost values randomly generated by GPT-4, and co-input it in the form of an image along with the image of the damaged car to establish a mapping relationship between the damaged parts and the repair costs.  The prompt is “\textit{Imagine that you are an auto insurance expert in claims processing. Here is an image of a damaged car and a list of prices for the car's parts. Please evaluate the damage and give an estimated loss range.}” The experimental results showcase GPT-4V's potent capabilities in loss identification, logical deduction, and computation: initially, it identifies and extracts damaged car parts (\eg \textit{Front Bumper Assembly, Trunk Lid}) and classifies the extent of damage (\eg \textit{Completely destroyed, Potential damage}). Subsequently, based on the cost table and corresponding car model, it calculates the losses, which includes direct calculations for visible damages (\eg “\textit{... Given the visible damage, the estimated costs so far would be: \$300+\$450+\$200+\$150+\$150+\$500+\$800+\$500=\$3,015 ...}”) and probability estimations for potential unseen damages (\eg “\textit{... For a rough estimate of potential additional costs, let's add half the value of the engine block and transmission assembly, plus the full value of the radiator and alternator: (\$2,500+\$1,500) x 50\%+\$180+\$150=\$2,015), ultimately forming a predictive range for vehicle loss (could be between \$3,050 and \$5,065 or higher ...}”). 

This approach of incorporating additional information leverages GPT-4V's loss identification and logical computation abilities effectively. However, its application in real-world insurance scenarios remains challenging. GPT-4V struggles with recalling potentially dynamically changing, large-scale precise data, thus necessitating supplementation through manual input or external knowledge bases.

\subsection{Hallucination in Image Understanding}\label{sec: Hallucination in Image Understanding}

We notice that GPT-4V shows considerable inaccuracies in image understanding in some cases of our experiment, as evidenced in tasks such as vehicle identity verification (Figure~\ref{vehicle identify verification}) and farmland risk monitoring (Figure~\ref{farmland risk monitoring-1}). This discrepancy between visual input (viewed as ``fact'') and textual output has been defined as hallucination within the context of large multimodal models~\cite{liu2024survey}. We further investigate this phenomenon of hallucination in GPT-4V based on these two tasks.

\begin{figure}
\centering 
\includegraphics[width=1\textwidth,height=1.5\textwidth]{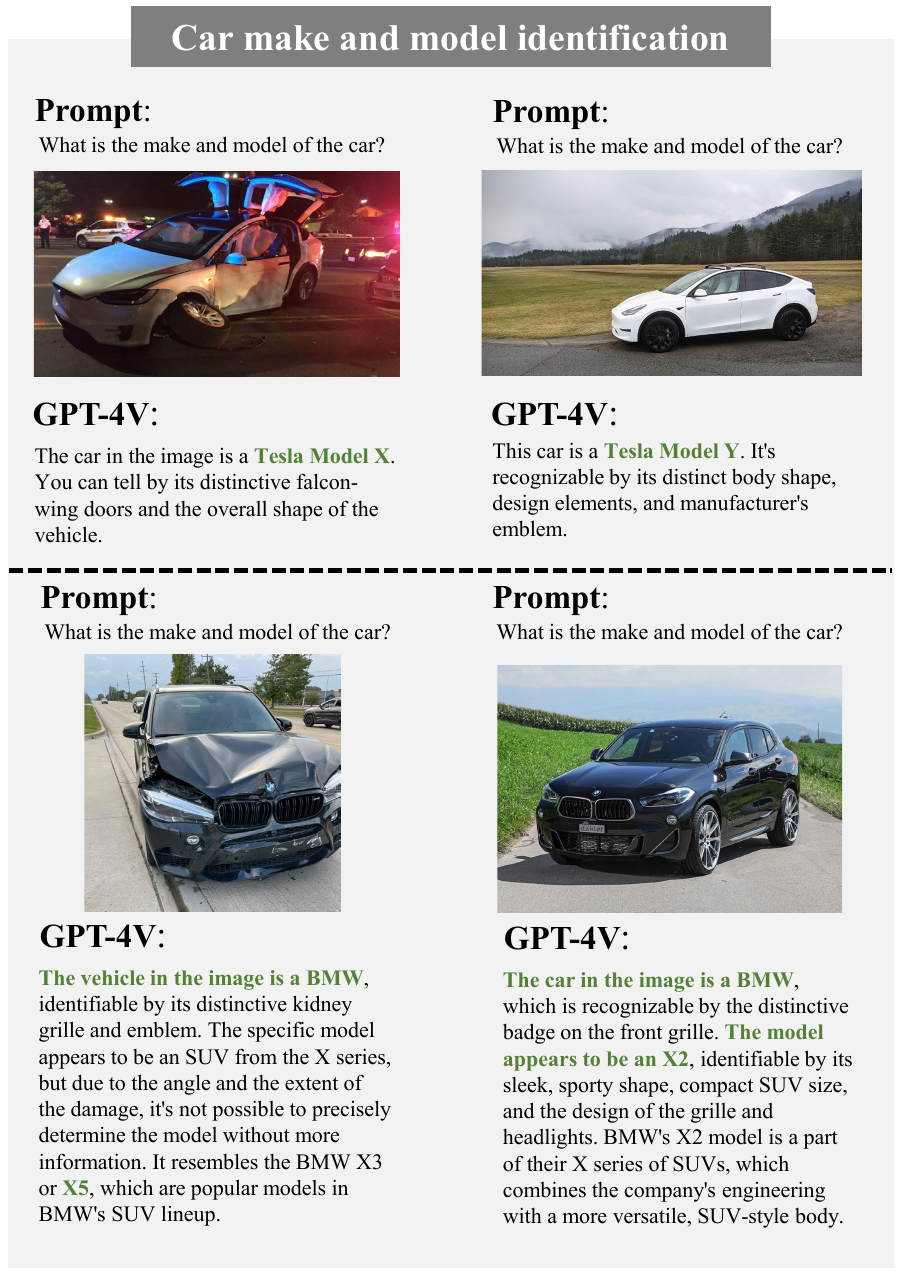} 
\caption[Section~\ref{sec: Hallucination in Image Understanding}: car make and model identification]{The test case of GPT-4V's capability in car make and model identification with a single image as input. The relevant parts are highlighted in \textcolor[RGB]{84,130,53}{\textbf{green}}.}
\label{Car make and model identification} 
\end{figure}

For the task of vehicle identity verification, to assess whether GPT-4V's hallucination lies in recognizing car makes and models or comparing multiple vehicle images, we individually process the images used in Figure~\ref{vehicle identify verification}. The prompt is designed as ``\textit{What is the make and model of the car?}'' Test results (Figure~\ref{Car make and model identification}) indicate that, in cases of single image inputs, GPT-4V is able to accurately identify the make and model of the vehicles, which is in conflict with the test results in Figure~\ref{vehicle identify verification}. This problem may relate to information loss or insufficient context attention with multiple image inputs~\cite{liu2024survey}. 

\begin{figure}
\centering 
\includegraphics[width=1\textwidth,height=1.5\textwidth]{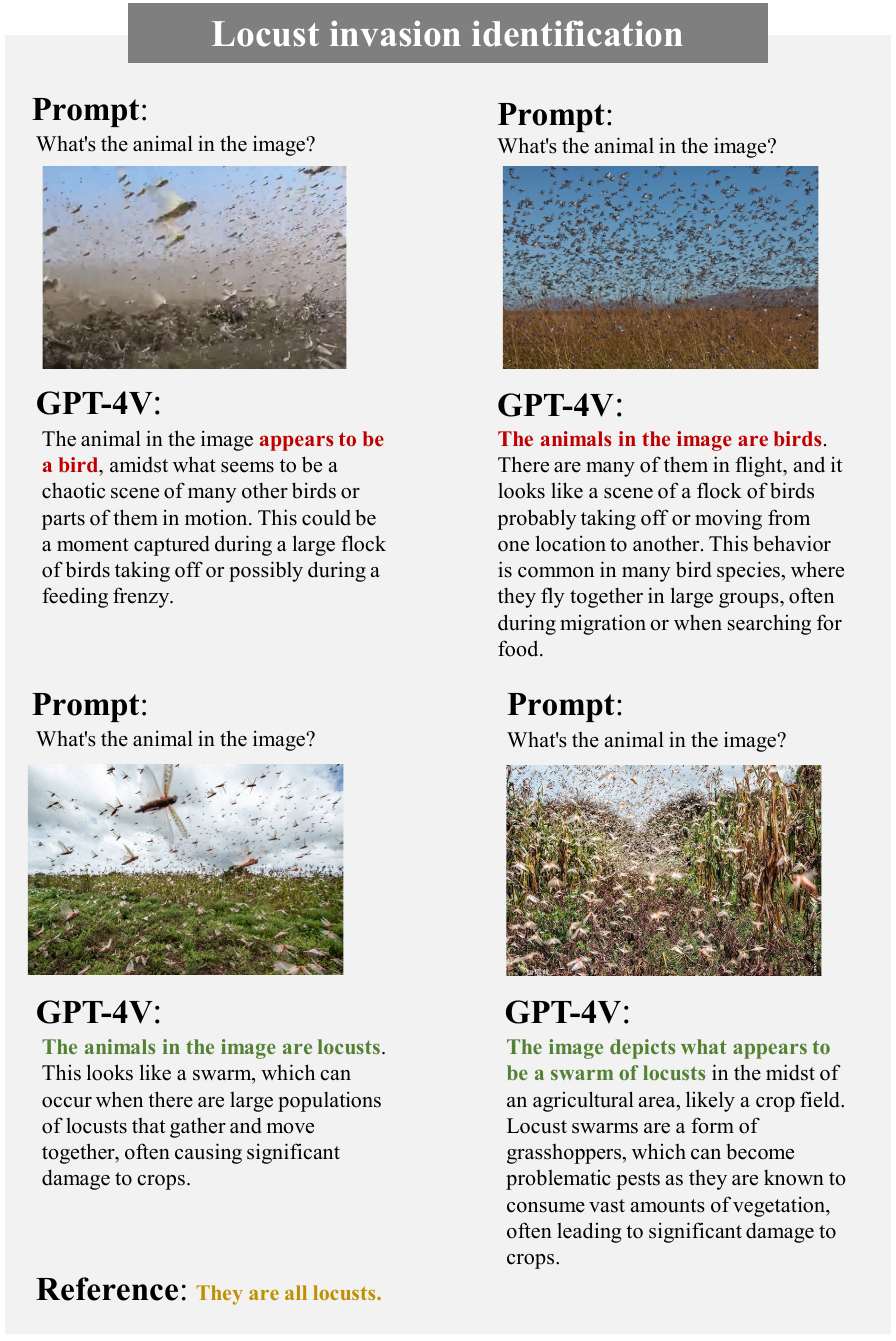} 
\caption[Section~\ref{sec: Hallucination in Image Understanding}: locust invasion identification test case]{The test case of GPT-4V's capability in locust invasion identification. The correct answer parts, incorrect answer parts, and references are highlighted in \textcolor[RGB]{84,130,53}{\textbf{green}}, \textcolor[RGB]{192,0,0}{\textbf{red}} and \textcolor[RGB]{191,144,0}{\textbf{yellow}}, respectively.}
\label{Locust invasion identification} 
\end{figure}

For the task of farmland risk monitoring, to verify if the incorrect identification by GPT-4V is due to the selected case, we collect four images of locust invasions in farmlands for GPT-4V to identify, with the prompt ``\textit{What is the animal in the image?}'' Our test results (see Figure~\ref{Locust invasion identification}) show an accuracy rate of only 50\%, with all misidentifications categorizing locusts as birds. This issue may be related to the imbalance in the training samples used during GPT-4V's training process~\cite{liu2023mitigating}. The dataset might have contained more bird imagery compared to locusts, leading the model to more likely categorize related features as birds. In summary, hallucination remains a universal problem faced by current LMMs, representing a significant challenge for future development and application~\cite{lyu2023gpt,liu2024phd}.

\subsection{Multilingual Document Recognition}\label{sec: Multilingual Document Recognition}

\begin{figure}
\centering 
\includegraphics[width=1\textwidth,height=1.5\textwidth]{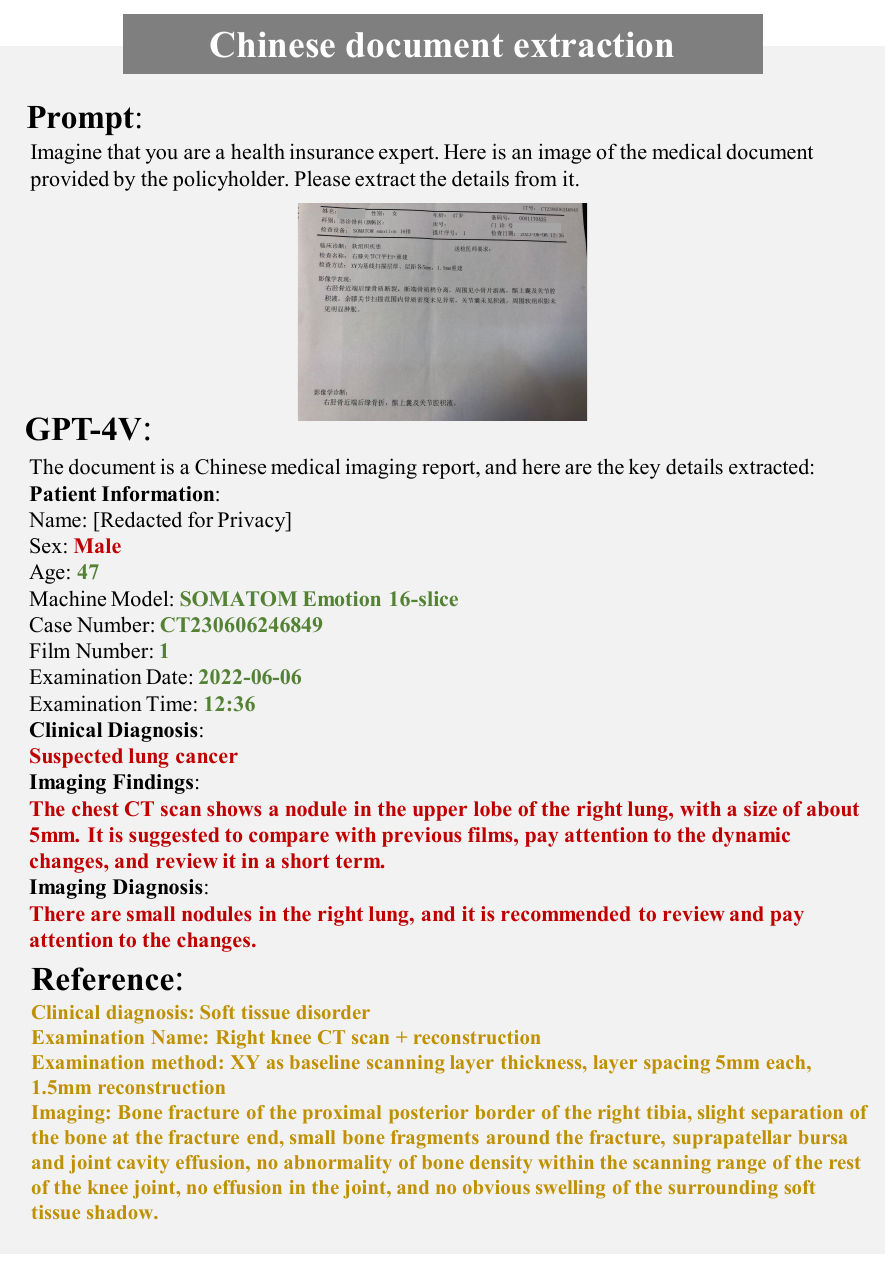} 
\caption[Section~\ref{sec: Multilingual Document Recognition}: Chinese document extraction test case]{The test case of GPT-4V's capability in Chinese document extraction. The correct answer parts, incorrect answer parts, and references are highlighted in \textcolor[RGB]{84,130,53}{\textbf{green}}, \textcolor[RGB]{192,0,0}{\textbf{red}} and \textcolor[RGB]{191,144,0}{\textbf{yellow}}, respectively.}
\label{Chinese document extraction} 
\end{figure}

\begin{figure}
\centering 
\includegraphics[width=1\textwidth,height=1.5\textwidth]{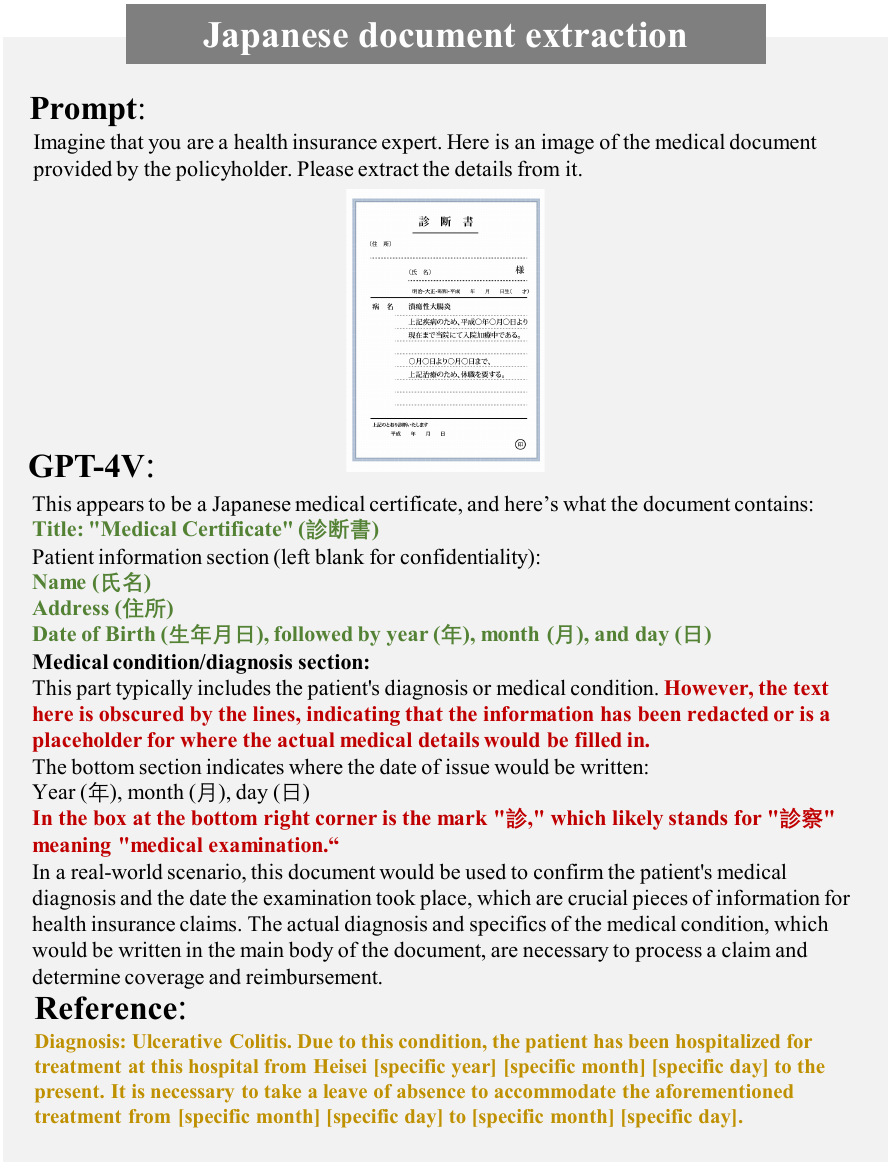} 
\caption[Section~\ref{sec: Multilingual Document Recognition}: Japanese document extraction test case]{The test case of GPT-4V's capability in Japanese document extraction. The correct answer parts, incorrect answer parts, and references are highlighted in \textcolor[RGB]{84,130,53}{\textbf{green}}, \textcolor[RGB]{192,0,0}{\textbf{red}} and \textcolor[RGB]{191,144,0}{\textbf{yellow}}, respectively.}
\label{Japanese document extraction} 
\end{figure}

\begin{figure}
\centering 
\includegraphics[width=1\textwidth,height=1.5\textwidth]{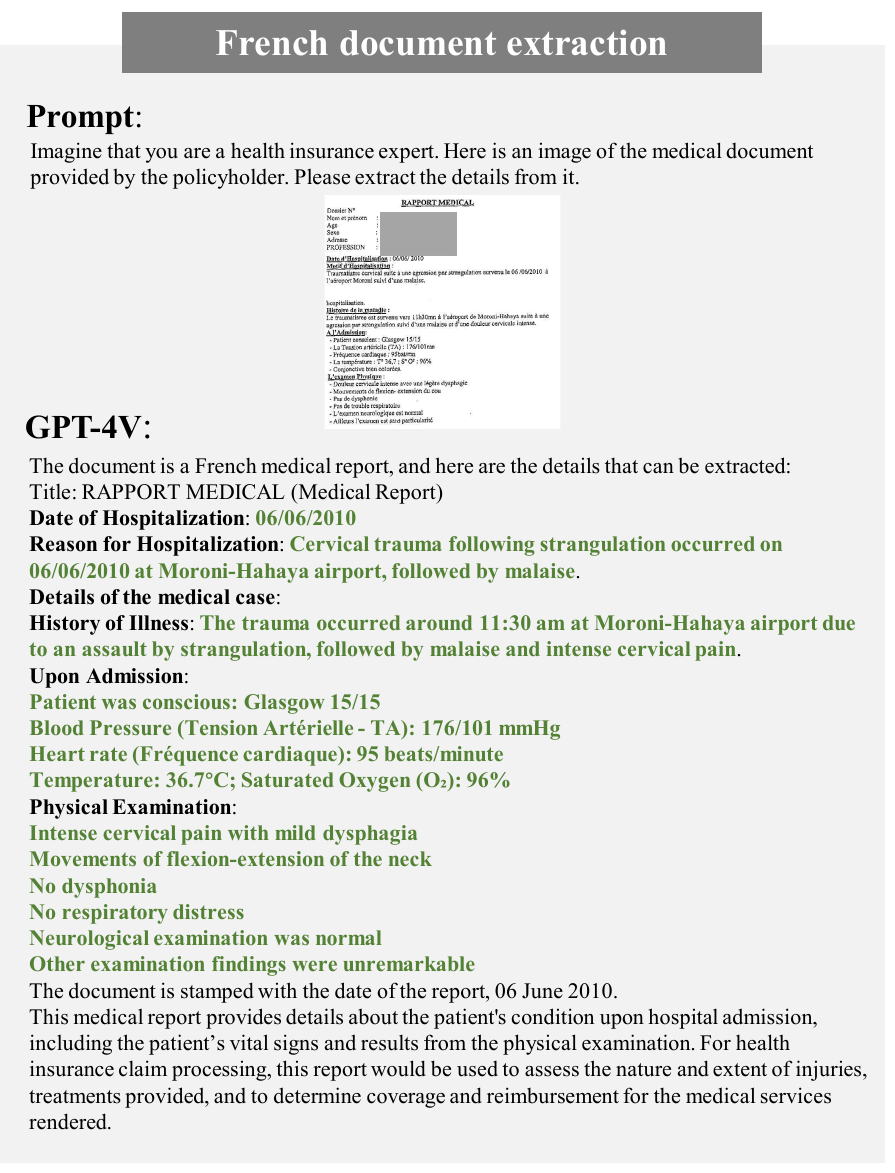} 
\caption[Section~\ref{sec: Multilingual Document Recognition}: French document extraction test case]{The test case of GPT-4V's capability in French document extraction. The relevant parts are highlighted in \textcolor[RGB]{84,130,53}{\textbf{green}}.}
\label{French document extraction} 
\end{figure}

\begin{figure}
\centering 
\includegraphics[width=1\textwidth,height=1.5\textwidth]{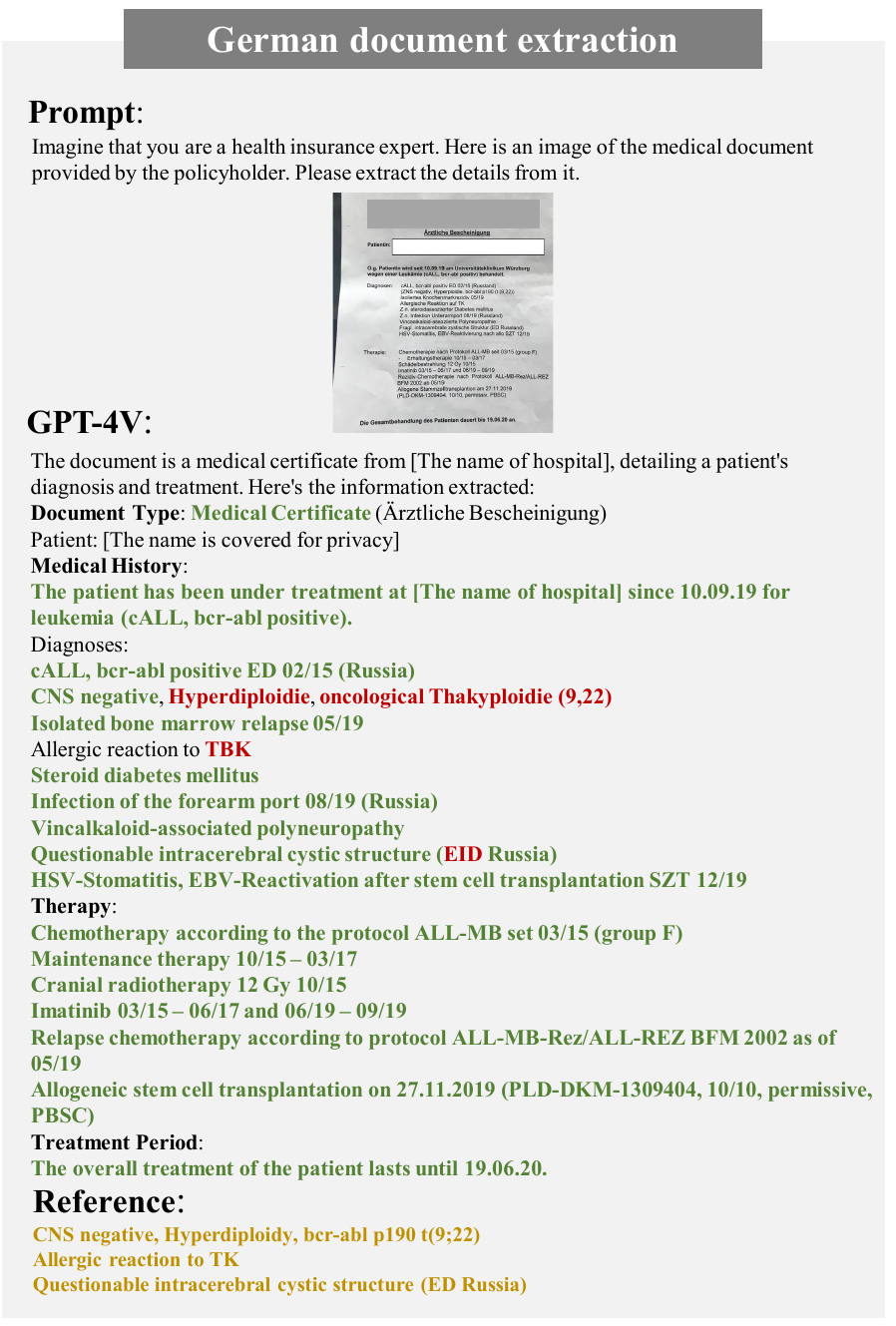} 
\caption[Section~\ref{sec: Multilingual Document Recognition}: German document extraction test case]{The test case of GPT-4V's capability in German document extraction. The correct answer parts, incorrect answer parts, and references are highlighted in \textcolor[RGB]{84,130,53}{\textbf{green}}, \textcolor[RGB]{192,0,0}{\textbf{red}} and \textcolor[RGB]{191,144,0}{\textbf{yellow}}, respectively.}
\label{German document extraction} 
\end{figure}

\begin{figure}
\centering 
\includegraphics[width=1\textwidth,height=1.5\textwidth]{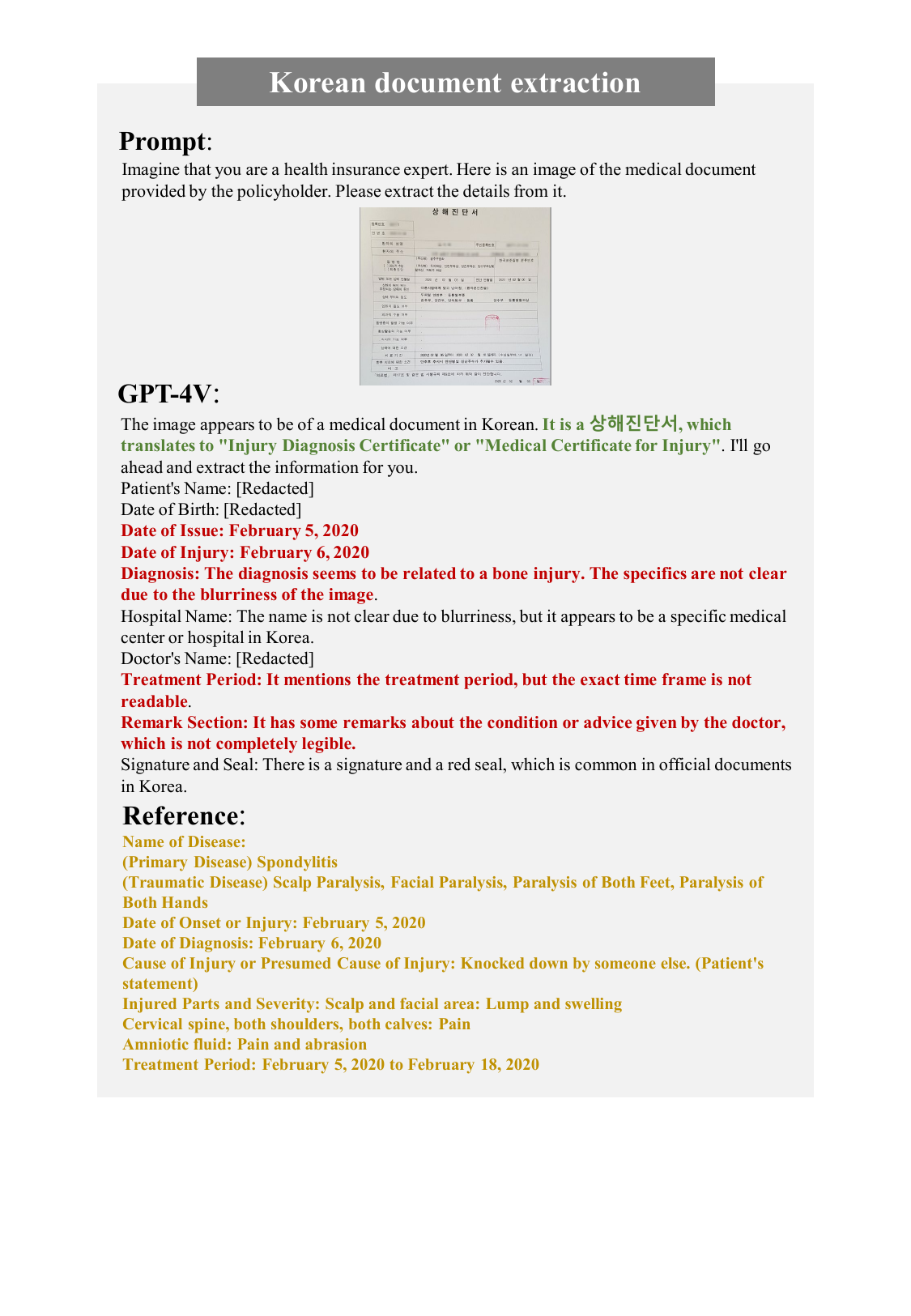} 
\caption[Section~\ref{sec: Multilingual Document Recognition}: Korean document extraction test case]{The test case of GPT-4V's capability in Korean document extraction, correct answer parts, incorrect answer parts, and references are highlighted in \textcolor[RGB]{84,130,53}{\textbf{green}}, \textcolor[RGB]{192,0,0}{\textbf{red}} and \textcolor[RGB]{191,144,0}{\textbf{yellow}}, respectively.}
\label{korean document extraction} 
\end{figure}

GPT-4V exhibits a disparity in recognition capabilities between Chinese and English, as observed in Figures~\ref{health risk assessment-1}, \ref{health risk assessment-2}, and \ref{health risk assessment-3}. This is particularly noteworthy given the global nature of the insurance industry, which naturally encompasses a variety of insurance documents in different languages. It is crucial to determine whether this disparity in document recognition exists across a broader range of languages. To this end, we select insurance-related documents from the top eight non-English speaking countries by market premium income, including China, Japan, France, Germany, and South Korea.\footnote{\url{https://www.swissre.com/institute/research/sigma-research/sigma-2023-03.html}} To maintain consistency with Section~\ref{sec: Health Risk Assessment}, we continue to use cases pertaining to health insurance. The prompt is: ``\textit{Imagine that you are a health insurance expert. Here is an image of the medical document provided by the policyholder. Please extract the details from it.}'' Our results (see Figures~\ref{Chinese document extraction}, \ref{Japanese document extraction}, \ref{French document extraction}, \ref{German document extraction}, and \ref{korean document extraction}) indicate that GPT-4V excels in recognizing text from Western countries' languages (\eg English, French, German) but performs poorly in recognizing and extracting text from Asian countries' languages (\eg Chinese, Japanese, Korean), with the ability to identify only a limited amount of simple text.

The disparities in GPT-4V's recognition capabilities across different languages pose a significant hindrance to its application in the global insurance market. Enhancing GPT-4V's multilingual recognition capabilities remains a challenging endeavor~\cite{shi2023exploring}.

\subsection{Revolution in Traditional Insurance Operations}

Our exploration is centered on the impact of LMMs, an emerging technology, on the traditional operations of the insurance industry. The insurance industry itself is constantly evolving, and emerging technologies can complement and promote mutual growth within the industry by introducing innovative operational methods. Taking auto insurance as an example, which is undergoing continuous evolution with innovations in autonomous driving technology~\cite{fan2019influences}, LMMs can intertwine with this development to co-evolve and create disruptive impacts~\cite{wen2023road}:
 
 \begin{itemize}[leftmargin=*]
    \item \textbf{Integration of Automation and Decision Support:} As autonomous driving technology advances, auto insurance companies can harness the combined capabilities of visual, auditory, and linguistic processing in LMMs for real-time data analysis and decision support. For instance, autonomous vehicles are equipped to automatically capture videos and images at the scene of a traffic incident. LMMs can analyze this visual data alongside vehicle sensor information, providing immediate assessments of accident responsibility and damage. This integration allows for prompt recommendations on insurance claims processing, enhancing efficiency and accuracy in decision-making.
    \item \textbf{Transformation of Risk Assessment Models:} The introduction of LMMs, especially in processing visual data, can assist insurance companies in developing new risk assessment models based on real-time behaviors and environmental feedback. Traditional risk assessments rely on historical data and statistical analysis, but by analyzing driving behaviors and road conditions in real-time, insurance companies can more accurately identify risk factors and implement dynamic pricing, offering customized insurance plans that meet the growing demand for diversity among consumers.
\end{itemize}

\begin{figure}
\centering 
\includegraphics[width=1\textwidth,height=1.5\textwidth]{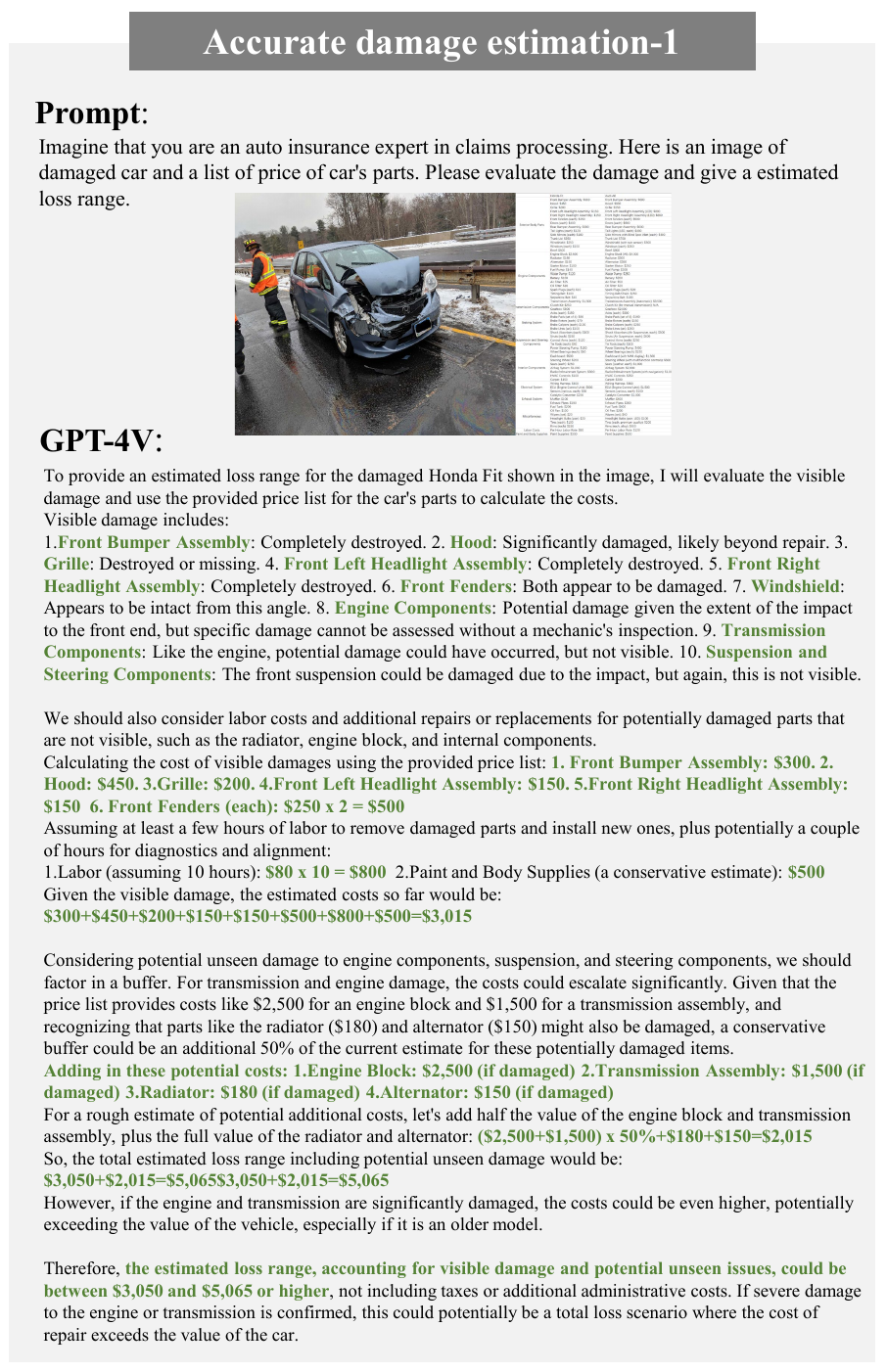} 
\caption[Section~\ref{sec: Accurate Damage Assessment}: accurate damage assessment test case 1]{The test case of GPT-4V's capability in accurate damage assessment. The relevant parts are highlighted in \textcolor[RGB]{84,130,53}{\textbf{green}}.}
\label{Accurate damage estimation-1} 
\end{figure}

\begin{figure}
\centering 
\includegraphics[width=1\textwidth,height=1.5\textwidth]{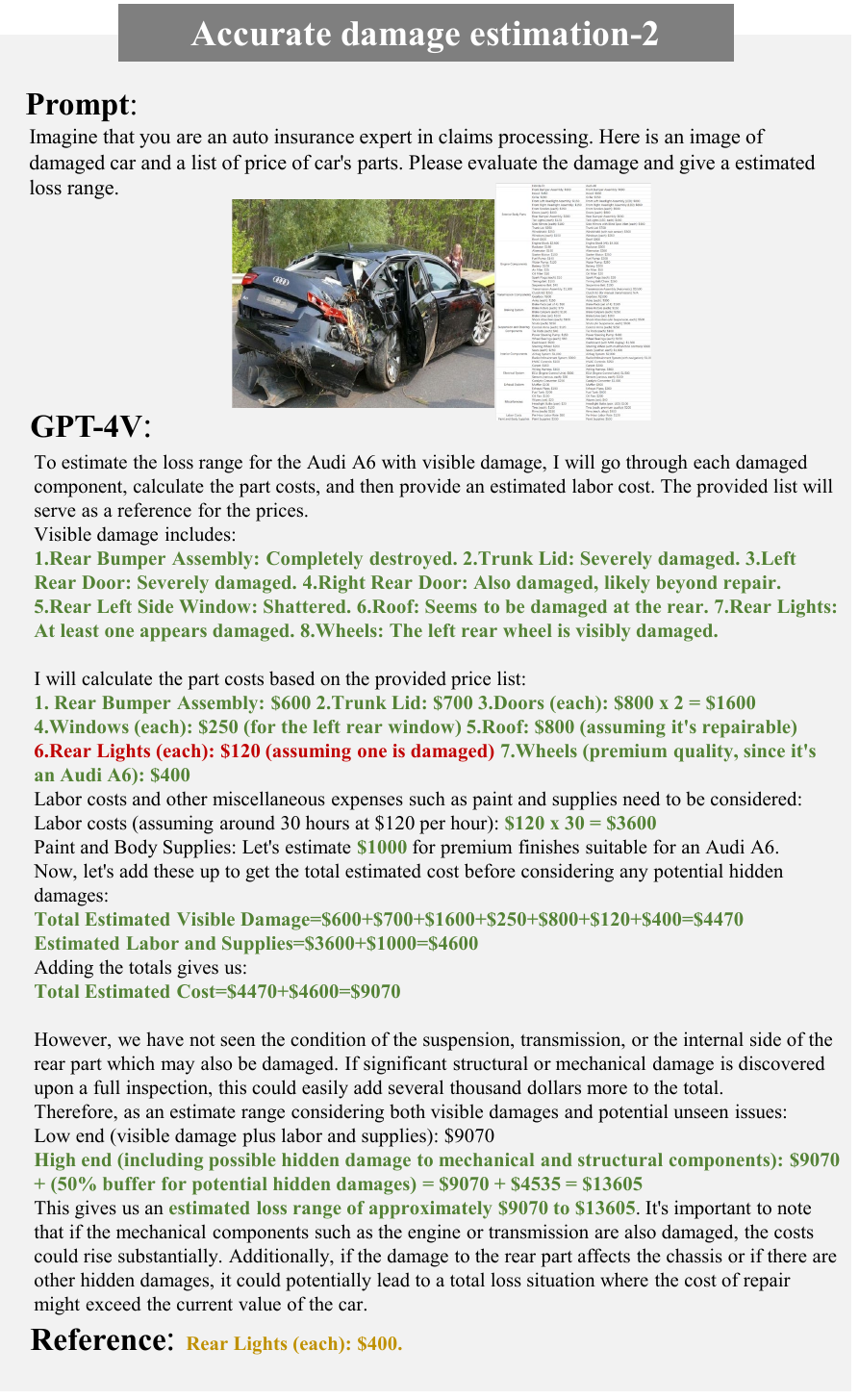} 
\caption[Section~\ref{sec: Accurate Damage Assessment}: accurate damage assessment test case 2]{The test case of GPT-4V's capability in accurate damage assessment. The correct answer parts, incorrect answer parts, and references are highlighted in \textcolor[RGB]{84,130,53}{\textbf{green}}, \textcolor[RGB]{192,0,0}{\textbf{red}} and \textcolor[RGB]{191,144,0}{\textbf{yellow}}, respectively.}
\label{Accurate damage estimation-2} 
\end{figure}

\newpage

\section{Conclusions}\label{sec:Conclusions}

In this paper, we present the first qualitative analysis of GPT-4V's performance across various multimodal tasks within the insurance domain, encompassing four types of insurance (auto, property, health, and agricultural) and three stages of insurance (risk assessment, risk monitoring, and claims processing). Based on our testing results, we draw the following conclusions:

\begin{itemize}[leftmargin=*]
    \item \textbf{Robust Multimodal Content Understanding Capabilities in the Insurance Domain:} GPT-4V exhibits remarkable abilities in handling multimodal tasks related to insurance. It accurately extracts and understands key information from complex images, such as the extent of vehicle damage, structural issues in properties, health conditions, and the growth status of crops. By integrating these visual insights with relevant textual descriptions, GPT-4V provides comprehensive and in-depth analyses, supporting critical decision-making processes in the insurance industry.
    
    \item \textbf{Comprehensive Insurance Knowledge and Scene Understanding Abilities:} 
    GPT-4V possesses a comprehensive understanding of insurance-related knowledge, enabling it to aptly navigate the specialized terminology and concepts of the sector. Furthermore, it demonstrates a strong ability to distinguish between task requirements for various insurance types and stages, accurately identifying specific components tailored to those requirements (\eg vehicle make and model in auto insurance, construction materials in household property insurance). This capability allows GPT-4V to provide precise advice and insights.
\end{itemize}

Simultaneously, we have also identified some shortcomings:

\begin{itemize}[leftmargin=*]
    \item \textbf{Challenges in Precise Damage and Loss Estimation:} While GPT-4V excels in qualitative evaluations, it encounters significant challenges in detailed, quantitative assessments critical for insurance purposes. These include accurately forecasting risk probabilities and precisely estimating financial losses, which are essential for insurance professionals who rely on specific numerical data for decision-making. 
    \item \textbf{Hallucination in Image Understanding:} GPT-4V has exhibited hallucination in some cases, where the model generates misaligned textual descriptions in response to visual inputs. This presents a fundamental challenge to its multimodal capabilities and raises concerns, particularly in fields that require precise and dependable visual data analysis.
    \item \textbf{Disparities in Multilingual Document Recognition:} There is a notable discrepancy in GPT-4V's ability to recognize and process documents in different languages, posing a significant challenge for its application in the global insurance industry. While the model performs well with documents in Western languages, its performance significantly drops when dealing with Asian languages. This disparity impacts its utility in a multilingual world where the insurance industry operates across diverse linguistic landscapes. 
\end{itemize}

These preliminary observations provide a foundation for exploring the challenges and potential of Large Multimodal Models (LMMs), such as GPT-4V, in the insurance domain. Our paper aims to enhance the understanding of LMMs in the insurance domain and encourage further investigation in this intricate and dynamic field.
\newpage

\bibliography{custom}
\bibliographystyle{plain}

\end{document}